\pdfoutput=1
\PassOptionsToPackage{unicode}{hyperref}
\PassOptionsToPackage{hyphens}{url}
\PassOptionsToPackage{dvipsnames,svgnames,x11names}{xcolor}
\documentclass[
  10pt]{article}
\usepackage{xcolor}
\usepackage[letterpaper,left=1.4in,right=1.4in,top=1in,bottom=1in]{geometry}
\usepackage{amsmath,amssymb}
\usepackage{iftex}
\ifPDFTeX
  \usepackage[T1]{fontenc}
  \usepackage[utf8]{inputenc}
  \usepackage{textcomp} 
\else 
\fi
\usepackage[]{mathptmx}
\ifPDFTeX\else
\fi
\IfFileExists{upquote.sty}{\usepackage{upquote}}{}
\IfFileExists{microtype.sty}{
  \usepackage[]{microtype}
  \UseMicrotypeSet[protrusion]{basicmath} 
}{}
\makeatletter
\@ifundefined{KOMAClassName}{
  \IfFileExists{parskip.sty}{%
    \usepackage{parskip}
  }{
    \setlength{\parindent}{0pt}
    \setlength{\parskip}{6pt plus 2pt minus 1pt}}
}{
  \KOMAoptions{parskip=half}}
\makeatother
\usepackage{longtable,booktabs,array}
\usepackage{calc} 
\usepackage{etoolbox}
\makeatletter
\patchcmd\longtable{\par}{\if@noskipsec\mbox{}\fi\par}{}{}
\makeatother
\IfFileExists{footnotehyper.sty}{\usepackage{footnotehyper}}{\usepackage{footnote}}
\makesavenoteenv{longtable}
\usepackage{graphicx}
\makeatletter
\newsavebox\pandoc@box
\newcommand*\pandocbounded[1]{
  \sbox\pandoc@box{#1}%
  \Gscale@div\@tempa{\textheight}{\dimexpr\ht\pandoc@box+\dp\pandoc@box\relax}%
  \Gscale@div\@tempb{\linewidth}{\wd\pandoc@box}%
  \ifdim\@tempb\p@<\@tempa\p@\let\@tempa\@tempb\fi
  \ifdim\@tempa\p@<\p@\scalebox{\@tempa}{\usebox\pandoc@box}%
  \else\usebox{\pandoc@box}%
  \fi%
}
\def\fps@figure{htbp}
\makeatother
\providecommand{\tightlist}{%
  \setlength{\itemsep}{0pt}\setlength{\parskip}{0pt}}
\usepackage[scaled=0.92]{helvet}
\usepackage{courier}
\usepackage{microtype}
\usepackage{titlesec}
\titleformat{\section}{\large\bfseries}{}{0em}{}
\titleformat{\subsection}{\normalsize\bfseries}{}{0em}{}
\titleformat{\subsubsection}{\normalsize\bfseries\itshape}{}{0em}{}
\titlespacing*{\section}{0pt}{1.6ex plus 1ex minus .2ex}{0.9ex plus .2ex}
\titlespacing*{\subsection}{0pt}{1.2ex plus .8ex minus .2ex}{0.6ex plus .2ex}
\usepackage{caption}
\usepackage{float}
\usepackage{etoolbox}
\AtBeginEnvironment{longtable}{\small\setlength{\tabcolsep}{4pt}}
\usepackage{booktabs}
\usepackage[numbers,sort&compress]{natbib}
\usepackage{titling}
\pretitle{\begin{center}\LARGE\bfseries}
\posttitle{\par\end{center}\vskip 0.4em}
\preauthor{\begin{center}\large}
\postauthor{\par\end{center}}
\predate{}\postdate{}
\usepackage{bookmark}
\IfFileExists{xurl.sty}{\usepackage{xurl}}{} 
\makeatletter
\@ifundefined{xmpquote}{}{}
\makeatother
\hypersetup{
  colorlinks=true,
  linkcolor={RoyalBlue},
  filecolor={Maroon},
  citecolor={RoyalBlue},
  urlcolor={RoyalBlue},
  pdfcreator={LaTeX via pandoc}}

\author{}
\date{}

\begin{document}

\thispagestyle{empty}
\begin{center}
{\LARGE\bfseries Value-Monotonicity Matters: A Concordance Loss for Deep Survival Prediction\par}
\vskip 0.7em
{\large Meixu Chen\textsuperscript{1,2,*}\quad Kai Wang\textsuperscript{1}\quad Jing Wang\textsuperscript{2}\par}
\vskip 0.4em
{\small \textsuperscript{1}Department of Radiation Oncology, University of Colorado School of Medicine, Aurora, CO, USA\par}
{\small \textsuperscript{2}Department of Radiation Oncology, University of Texas Southwestern Medical Center, Dallas, TX, USA\par}
\vskip 0.2em
{\small \textsuperscript{*}Corresponding author: \href{mailto:Meixu.Chen@CUAnschutz.edu}{Meixu.Chen@CUAnschutz.edu}\par}
\end{center}
\vskip 0.6em
\begin{abstract}
\noindent Deep survival models are evaluated almost exclusively by the concordance index (C-index), yet they are trained on likelihood objectives such as the Cox partial likelihood, discrete-time negative log-likelihood, and the likelihood term of DeepHit. This mismatch between the training objective and the evaluation metric is usually considered harmless, because the C-index can be recomputed on a validation set at every epoch to guide monitoring, early stopping, and model selection. In the setting that motivates this work, however, where a high-capacity encoder is trained end to end with a survival head on the small, heavily censored cohorts typical in oncology, that recomputation is too expensive to perform often, so the value of the loss itself becomes the signal these decisions rely on. We show that this reliance is unsafe for likelihood losses, and we propose a remedy. First, we prove that every strictly proper survival likelihood admits a direction in which the loss strictly decreases while the C-index is exactly unchanged, so the value of the loss decouples from the metric during training, most severely once the ranking has converged. Second, we study a sigmoid concordance loss (SCL) whose value equals one minus the C-index up to a temperature term, which makes lower loss correspond to higher C-index throughout optimization; the loss is architecture agnostic and, for linear models, reduces to a convex survival ranking support vector machine. We evaluate the loss across four different modalities spanning eighteen datasets under a single five-fold cross-validation protocol. The SCL matches the discrimination of the standard likelihood losses, being best or within approximately one standard deviation of the best method on the C-index, while being the only loss whose value tracks the C-index during training, with a rank correlation between the loss and the metric of approximately 0.96 to 0.99 against values that range from slightly negative to about 0.53 for the likelihood losses. Calibration measured by the integrated Brier score is comparable with the likelihood losses. The contribution of this work is not a new state of the art on discrimination but a loss whose value can be trusted as a proxy for the C-index during expensive end-to-end training. Code and preprocessing pipelines are released at \url{https://github.com/Meixu-Chen/sigmoid-concordance-loss} to support reproduction.

\smallskip\noindent\textbf{Keywords:} survival analysis, concordance index, loss function, deep learning, medical imaging, model selection.
\end{abstract}
\vskip 0.4em

\hypertarget{sec:1}{}\subsection{1. Introduction}\label{introduction}

Survival prediction estimates the time from a baseline observation to a
clinical event, such as death, recurrence, or progression, from patient
data that includes censored observations for whom the event has not yet
occurred. It underlies risk stratification, treatment selection, and
clinical trial design across oncology and beyond. In practice the
quantity a clinician acts on is rarely a calibrated survival curve; it
is a risk ordering that separates patients who are likely to experience
the event sooner from those who are likely to experience it later. This
is why the concordance index, introduced by Harrell et
al.~\cite{harrell1982} and refined for censored data by Uno et
al.~\cite{uno2011}, has become the near-universal measure of a survival
model. The concordance index is a rank statistic that reports how often
a model orders the risks of two comparable patients in agreement with
their observed outcomes, and it depends on the model only through the
ordering it induces.

Machine learning, and deep learning in particular, has moved survival
prediction from handcrafted prognostic scores toward models that learn
representations directly from high-dimensional inputs. DeepSurv
\cite{katzman2018} parameterized the Cox proportional hazards model
\cite{cox1972} with a neural network; DeepHit \cite{lee2018} modeled the
discrete-time event distribution directly and added a ranking term;
multi-task logistic regression \cite{yu2011}, realized as a neural
discrete-time model, and the logistic-hazard family
\cite{gensheimer2019,kvamme2019} predict interval hazards; and
continuous-time and parametric approaches such as SODEN \cite{tang2022}
and Deep Survival Machines \cite{nagpal2021} fit event distributions
with neural networks. In computational pathology, survival models attach
a prediction head to multiple-instance aggregators over foundation-model
features, as in PORPOISE \cite{chen2022porpoise}, MCAT
\cite{chen2021mcat}, and SurvPath \cite{jaume2024}. These methods differ
in architecture and in how they represent time, but they share a common
design choice at the level of the objective: they are trained by
maximizing a likelihood, and they are evaluated by the C-index.

The objective and the metric are therefore not the same kind of
quantity. A likelihood is a strictly proper scoring rule whose
population optimum is the true conditional event distribution, a single
point in distribution space. The C-index is a function of the ordering
alone and is invariant to any monotone rescaling of the risk score. The
standard justification for training on a likelihood and evaluating on
the C-index is that the two are correlated at the optimum, and that in
any case the C-index can be computed on a held-out set at each epoch to
monitor progress, stop early, and select the final checkpoint. That
justification is reasonable when evaluation is cheap. It becomes
unreliable in the expensive end-to-end training regime that motivates
this paper. When a high-capacity image encoder is trained end to end
together with a survival head, on the small and heavily censored cohorts
that are common in oncology, each training epoch and each full
validation pass is costly, so the C-index is computed only occasionally
and the loss becomes the operative signal for the decisions that
determine which model is deployed. If the value of the loss does not
move with the C-index, then early stopping, checkpoint selection, and
hyperparameter comparison are made on a signal that no longer reflects
the metric of interest, and they can silently degrade the deployed
model.

This paper hypothesizes that this shortcoming is structural for
likelihood losses and that it is avoidable. We make three contributions.

\begin{enumerate}
\def\labelenumi{\arabic{enumi}.}
\tightlist
\item
  We formalize the property that a survival loss must have for its value
  to be usable as a proxy for the C-index during training, which we call
  value-monotonicity, and we prove that the SCL has this property,
  because its value equals one minus the C-index up to a term that
  vanishes with the temperature.
\item
  We prove that this decoupling is unavoidable for likelihood losses:
  every strictly proper survival likelihood admits an explicit
  direction, always available through rescaling of the risk score, along
  which the loss strictly decreases while the C-index is exactly
  unchanged, so its value must decouple from the metric, and this
  decoupling is most pronounced late in training when the ranking has
  stabilized and the loss is trusted most.
\item
  We validate these claims across four modalities and eighteen datasets
  under one five-fold cross-validation protocol, and we show that the
  SCL matches the discrimination of the standard likelihood losses while
  being the only loss whose value stably tracks the C-index and supports
  near-zero-regret model selection, at comparable calibration.
\end{enumerate}

We are deliberate about the scope of the empirical claim. The SCL does
not dominate the field on the C-index; it matches it. The contribution
is a loss whose value can be trusted, which is the property the
expensive end-to-end imaging regime actually needs. All code,
preprocessing, and evaluation scripts are released so that the protocol
can be reproduced.

\hypertarget{sec:2}{}\subsection{2. Related work}\label{related-work}

\hypertarget{sec:2.1}{}\subsubsection{2.1 Loss functions for deep survival
analysis}\label{loss-functions-for-deep-survival-analysis}

Most deep survival models optimize a likelihood. DeepSurv
\cite{katzman2018} minimizes the negative Cox partial log-likelihood
over risk sets. Discrete-time models partition the time axis into
intervals and predict conditional hazards; multi-task logistic
regression \cite{yu2011} and its neural realisations, together with the
logistic-hazard and Nnet-survival formulations \cite{gensheimer2019},
minimize the corresponding interval negative log-likelihood. DeepHit
\cite{lee2018} models the discrete event distribution directly and
combines a likelihood term with an exponential ranking term. SODEN
\cite{tang2022} and Deep Survival Machines \cite{nagpal2021} fit
continuous-time or parametric likelihoods. TripleSurv \cite{zhang2024}
augments the training objective with a time-adaptive pairwise ranking
term. In computational pathology, PORPOISE \cite{chen2022porpoise}, MCAT
\cite{chen2021mcat}, and SurvPath \cite{jaume2024} attach a
discrete-time negative log-likelihood head to multiple-instance
aggregators over features from pathology foundation models such as UNI
\cite{chen2024uni}, CONCH \cite{lu2024conch}, and TITAN
\cite{ding2024titan}. Across this body of work the architecture and the
time representation vary, but the objective that is optimized is a
likelihood, while the reported measure of quality is a rank statistic.
Our contribution is orthogonal to the architecture and concerns the loss
alone: we hold the backbone fixed and change only the objective.

\hypertarget{sec:2.2}{}\subsubsection{2.2 Concordance surrogates and their
consistency}\label{concordance-surrogates-and-their-consistency}

Because the C-index is not differentiable, a body of work replaces it
with smooth pairwise surrogates. Steck et al.~\cite{steck2008}
introduced smooth bounds on the concordance index for survival ranking.
The statistical properties of pairwise surrogates have been analyzed
through the lens of consistency: Gao and Zhou \cite{gao2015}
characterized which pairwise surrogates are consistent for the area
under the receiver operating characteristic curve, showing that the
squared hinge, the logistic, and the sigmoid surrogates are consistent
while the plain hinge is not, and Elgui et al.~\cite{elgui2023}
established consistency results specific to the concordance index. The
general theory of surrogate losses and their calibration was developed
by Bartlett et al.~\cite{bartlett2006} and Steinwart
\cite{steinwart2007}. All of these results concern the minimizer of the
surrogate, that is, the guarantee that optimizing the surrogate far
enough recovers the optimum of the metric. Our claim is different and
stronger. We are concerned not only with where the surrogate is
minimized but with how its value behaves along the entire optimization
trajectory, because that is what determines whether the loss can be
trusted before convergence. We show that the sigmoid surrogate has a
value that tracks the metric throughout training, and that likelihood
losses, although they may be consistent, do not have this
trajectory-level property. To our knowledge, both the trajectory-level
property and the accompanying negative result for likelihood losses are
new, as is their use as a criterion for loss design.

\hypertarget{sec:2.3}{}\subsubsection{2.3 Training under a mismatch between loss and evaluation
metric}\label{training-under-a-mismatch-between-loss-and-evaluation-metric}

The gap between a differentiable training loss and a non-differentiable
evaluation metric is a recurring theme in learning to rank and in
metric-aware training. Surrogates for ranking metrics such as the
normalized discounted cumulative gain and the area under the curve have
been studied extensively, from RankNet \cite{burges2005} to general
frameworks for the direct optimization of information-retrieval measures
\cite{qin2010}. Survival prediction is a particularly clean instance of
this mismatch, because the evaluation metric is a rank statistic and the
standard losses are proper scoring rules, and because the practical cost
of evaluation in end-to-end imaging makes the trajectory-level behavior
of the loss consequential rather than merely aesthetic. We cast
survival-loss design in this frame and quantify the mismatch with
standard statistics, namely the rank correlation between the loss and
the metric along training and the regret incurred by selecting a model
on the loss, rather than introducing a new metric that would itself
require justification.

\hypertarget{sec:2.4}{}\subsubsection{2.4 Survival prediction across imaging
modalities}\label{survival-prediction-across-imaging-modalities}

Deep survival prediction has been applied to computed tomography and
positron emission tomography, most prominently in the head-and-neck
outcome-prediction task of the HECKTOR challenge
\cite{saeed2025,saeed2026}, to magnetic resonance imaging of brain
tumors in cohorts such as UCSF-PDGM \cite{calabrese2022} and UPENN-GBM
\cite{bakas2022}, and to pathology whole-slide images in the
multiple-instance survival models cited above. These studies typically
pair a modality-specific architecture with a likelihood loss and report
the C-index. Our experiments deliberately reuse standard, unremarkable
architectures for each modality and vary only the loss, so that any
difference in the loss-metric relationship is attributable to the
objective rather than to the network. This design reflects the goal of
the paper, which is to study and evaluate the loss function rather than
to advance any single architecture.

\hypertarget{sec:3}{}\subsection{3. Method}\label{method}

\hypertarget{sec:3.1}{}\subsubsection{3.1 Problem formulation}\label{problem-formulation}

We observe a right-censored survival dataset
\(\{(x_n, t_n, \delta_n)\}_{n=1}^N\), where \(x_n \in \mathcal{X}\) is a
covariate that may be a tabular vector, a three-dimensional image
volume, or a slide-level feature vector, \(t_n > 0\) is the observed
time, and \(\delta_n \in \{0,1\}\) indicates whether the event was
observed (\(\delta_n = 1\)) or the observation was censored
(\(\delta_n = 0\)). A model produces a scalar risk score
\(f(x) \in \mathbb{R}\), with a higher score denoting higher risk and
therefore shorter expected survival. A pair of subjects \((i,j)\) is
called comparable if the earlier time is an observed event, that is
\(t_i < t_j\) and \(\delta_i = 1\), because only then is their relative
ordering known. The concordance index is the probability that the model
orders a comparable pair correctly, \[
C(f) = \Pr\big( f(x_i) > f(x_j) \mid t_i < t_j,\ \delta_i = 1 \big),
\] with ties counted as one half. Harrell's and Uno's versions of the
concordance index differ only in how comparable pairs are weighted, and
both depend on the model through the ordering of \(f\) alone.

\hypertarget{sec:3.2}{}\subsubsection{3.2 Value-monotonicity: a training-time requirement on
the
loss}\label{value-monotonicity-a-training-time-requirement-on-the-loss}

Fisher consistency asks that the minimizer of a loss coincide with the
maximizer of the C-index. This is a statement about the optimum. When
the loss is used to monitor training, to stop early, and to select a
checkpoint, the relevant question is different: does the value of the
loss move with the C-index at every point of the optimization
trajectory, and not only at convergence. We formalize this as follows. A
loss \(L\) is value-monotone in the C-index if \(L(f)\) is a decreasing
function of \(C(f)\), so that a lower loss implies a higher C-index
throughout training rather than merely in the limit. Value-monotonicity
is the property that makes the loss a trustworthy stand-in for the
metric during training. The remainder of this section constructs a loss
with this property in \hyperlink{sec:3.3}{Section 3.3}, shows in \hyperlink{sec:3.5}{Section 3.5} that strictly
proper likelihood losses cannot have it, and defines in \hyperlink{sec:3.7}{Section 3.7} how
we measure the extent to which a loss has it in practice.

\hypertarget{sec:3.3}{}\subsubsection{3.3 The sigmoid concordance loss
(SCL)}\label{the-sigmoid-concordance-loss-scl}

For a minibatch, let \(\mathcal{P}\) denote its set of comparable pairs.
The proposed objective, which we refer to throughout as the SCL, is the
mean over comparable pairs of a smoothed indicator of misordering, \[
L_\tau(f) = \frac{1}{|\mathcal{P}|} \sum_{(i,j) \in \mathcal{P}} \sigma\!\left( -\frac{f(x_i) - f(x_j)}{\tau} \right), \qquad \sigma(u) = \frac{1}{1+e^{-u}},
\] with a temperature \(\tau > 0\), which we fix at \(\tau = 0.1\) in
all experiments rather than tuning it per dataset (Supplementary
Material, Section C). Two properties motivate this choice. First, the
loss is a smooth surrogate for the misordering rate. As the temperature
tends to zero, the sigmoid approaches the indicator of a wrong ordering,
and the value of the loss approaches one minus the C-index, as stated
formally in Theorem 1 of Appendix A. The value of the loss is therefore
an estimate of one minus the C-index, and not merely a quantity that
shares its optimum. Second, the loss depends on the score differences
only through the smoothed sign of the difference, so it is
asymptotically invariant to the magnitudes of the scores. This is the
reason we use the sigmoid rather than a margin-based surrogate such as
the squared hinge. A margin-based surrogate penalizes the size of the
score differences and therefore changes value when the scores are
rescaled without changing their order, which breaks the correspondence
between the value of the loss and the C-index. The consequence is
visible in the experiments, where the sigmoid surrogate attains a
loss-metric rank correlation close to 0.98 while the squared-hinge
surrogate attains roughly 0.4 at comparable discrimination. The loss is
architecture agnostic and is attached to any backbone that emits a
scalar risk score, which is what allows us to hold the architecture
fixed and vary only the objective.

\hypertarget{sec:3.4}{}\subsubsection{3.4 A convex linear special
case}\label{a-convex-linear-special-case}

For a linear risk model \(f(x) = w^\top x\), the squared-margin
relaxation of the SCL is convex in \(w\) and defines a survival ranking
support vector machine that can be solved to global optimality, for
example with a quasi-Newton method. This gives a convex classical
counterpart to the deep and stochastic version of the loss, analogous to
the way the linear Cox model underlies DeepSurv, and it is useful when a
globally optimal linear baseline is required. We implement this convex
linear counterpart and evaluate it under the same pooled five-fold
protocol on the ten tabular datasets. It attains a mean Harrell C-index
of 0.792, which matches the deep SCL at 0.785 and confirms that the
convex classical version recovers the discrimination of the deep
stochastic one rather than trading it away for convexity. The
per-dataset comparison of the convex linear counterpart and the deep SCL
is given in \hyperlink{tab:S3}{Table S3} of the Supplementary Material.

\hypertarget{sec:3.5}{}\subsubsection{3.5 Why strictly proper likelihood losses provably
decouple from the
C-index}\label{why-strictly-proper-likelihood-losses-provably-decouple-from-the-c-index}

All of the theoretical results rest on one structural fact, stated as
Lemma 1 in Appendix A: the C-index is invariant to strictly increasing
reparameterizations of the risk score, so that
\(C(\varphi \circ f) = C(f)\) for every strictly increasing \(\varphi\).
The level sets of the C-index are therefore the classes of scores that
share an ordering, and any direction that changes the values of a score
while preserving its order leaves the C-index unchanged to first order.

A survival likelihood loss, whether the Cox partial likelihood, a
discrete-time negative log-likelihood, or the likelihood term of
DeepHit, is a strictly proper scoring rule, so its population minimizer
is the true conditional event distribution, which is a single point
rather than an entire class of orderings. Proposition 2 in Appendix A
makes the consequence explicit for the always-available rescaling
direction. For any risk score \(f\) that is non-constant on some risk
set, the map \(\alpha \mapsto L_{\mathrm{cox}}(\alpha f)\) is strictly
convex, because its second derivative equals the mean over events of the
variance of \(f\) under the risk-set softmax, which is strictly
positive. Since rescaling preserves the ordering, the C-index is exactly
constant along this direction while the Cox likelihood strictly
decreases along it. Every model therefore admits an explicit direction
along which the likelihood decreases and the C-index does not move.
Theorem 2 generalizes this from the single rescaling direction to the
full group of monotone reparameterizations, showing that any loss that
is not invariant to that group has a gradient component along the
directions that leave the C-index unchanged. Corollary 2 draws the
training-time consequence: once the ranking has approximately converged,
the remaining likelihood gradient lies predominantly in these
order-preserving directions, so the loss continues to decrease while the
C-index moves only through finite-sample noise, and the value of the
loss becomes uninformative about the metric. We state this carefully as
uninformativeness rather than as systematic anti-correlation, because a
strongly negative rank correlation measured on a converged plateau is a
noise artefact rather than a genuine inversion. \hyperlink{fig:1}{Figure 1} illustrates the
mechanism on synthetic Cox data: as the risk score is rescaled, the
C-index remains constant while the Cox negative log-likelihood sweeps
across a wide range.

\begin{figure}
\hypertarget{fig:1}{}\centering
\includegraphics[width=0.58\linewidth,height=\textheight,keepaspectratio,alt={\hyperlink{fig:1}{Figure 1}. The decoupling mechanism on synthetic Cox data. Rescaling the risk score leaves the concordance index exactly constant while the Cox negative log-likelihood changes substantially, so the value of a likelihood loss can move without any change in the ranking that the concordance index measures.}]{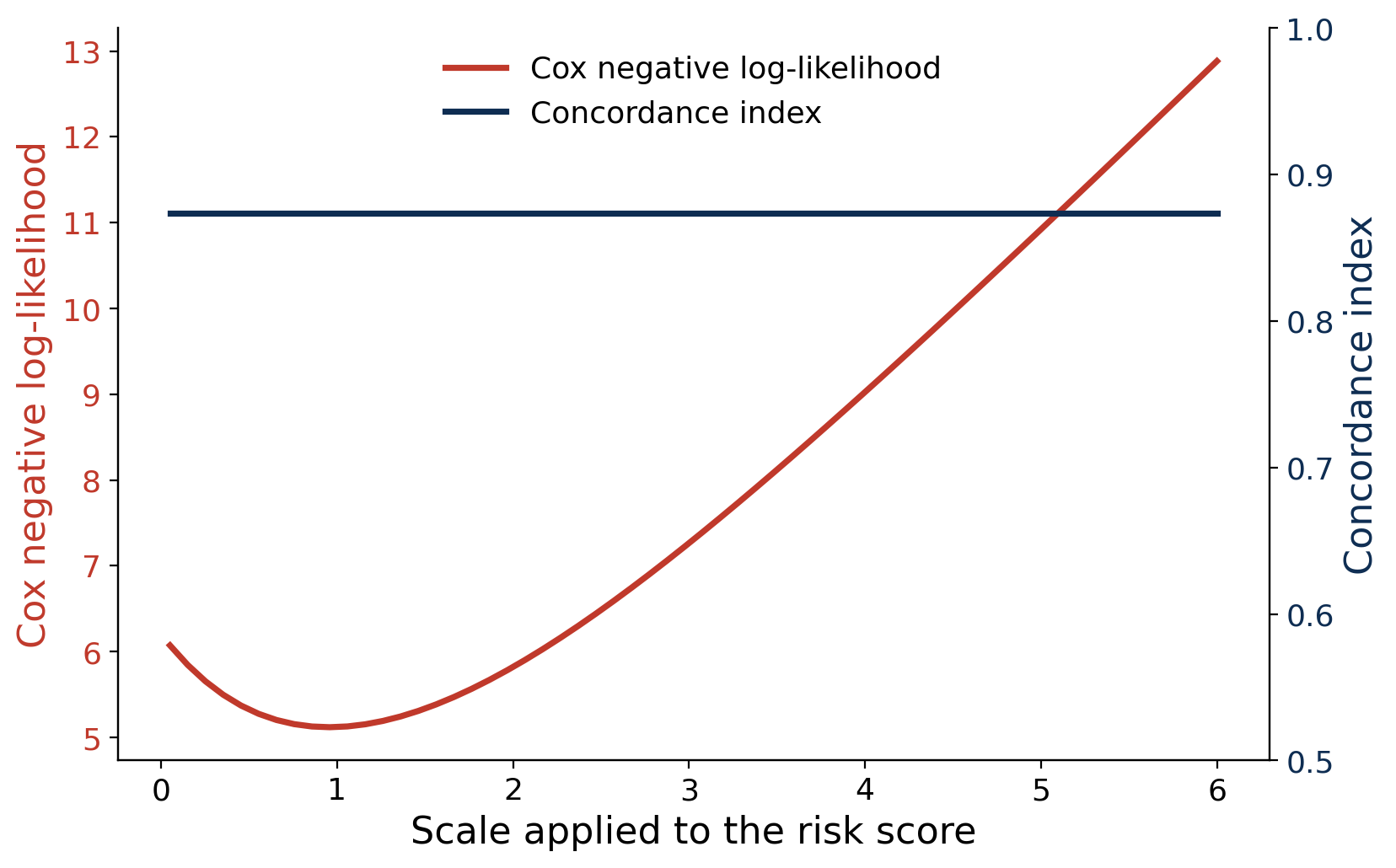}
\caption{Figure~1. The decoupling mechanism on synthetic Cox data.
Rescaling the risk score leaves the concordance index exactly constant
while the Cox negative log-likelihood changes substantially, so the
value of a likelihood loss can move without any change in the ranking
that the concordance index measures.}
\end{figure}

\hypertarget{sec:3.6}{}\subsubsection{3.6 A hybrid variant as a
diagnostic}\label{a-hybrid-variant-as-a-diagnostic}

To test the theory rather than to propose a competitor, we also study a
hybrid loss that adds a likelihood anchor to the SCL with weight
\(\lambda\), so that the objective is
\(L_\tau(f) + \lambda \, L_{\mathrm{lik}}(f)\). The theory predicts that
any positive anchor weight reintroduces the order-preserving gradient
direction of the likelihood term and therefore degrades
value-monotonicity. The hybrid is not intended as a stronger method, and
we report it only as a controlled ablation. Its expected behavior,
confirmed in \hyperlink{sec:5.4}{Section 5.4}, is that it matches neither the discrimination
advantage of the SCL nor its coupling, because adding the anchor lowers
the coupling without improving the C-index. We therefore interpret the
hybrid as evidence for the theory and treat the SCL as the proposed
method.

\hypertarget{sec:3.7}{}\subsubsection{3.7 Measuring loss-metric
coupling}\label{measuring-loss-metric-coupling}

Because the central claim concerns the relationship between the loss and
the metric during training, we quantify that relationship with two
standard, interpretable statistics rather than with a new metric. The
first is the Spearman rank correlation between the trajectory of the
negative validation loss and the trajectory of the validation C-index
over the later half of training, averaged over folds and seeds. We refer
to this as the loss-metric rank correlation. A value near one means that
as the validation loss falls the validation C-index rises in lockstep,
which is exactly the value-monotone behavior of \hyperlink{sec:3.2}{Section 3.2}. The second
statistic is the selection regret, which measures how good the loss is
as a signal for choosing the final model rather than how well the model
discriminates. Training produces one checkpoint per epoch, and exactly
one of them has to be chosen for deployment. That choice cannot be made
on the test concordance index, because the test outcomes are held out,
so it is made on a validation signal. Two signals are available. The
first is the validation concordance index, which is faithful to the
evaluation metric but has to be recomputed on the full validation set at
every epoch that is a selection candidate. The second is the validation
loss, which the training procedure has already produced at no additional
cost. The selection regret is the test concordance index of the
checkpoint chosen by the validation concordance index minus the test
concordance index of the checkpoint chosen by the validation loss. It is
therefore the amount of test discrimination that is forfeited by
selecting the model on the freely available loss instead of on the
costly metric. A selection regret near zero means the loss identifies a
checkpoint that is as good as the one the metric would have identified,
so the loss can be trusted for selection and the expensive per-epoch
evaluation of the concordance index can be skipped. A large selection
regret means the loss points to a worse checkpoint than the metric
would, so the loss is an unreliable selection signal. It is important
that the selection regret evaluates the loss as a selection signal and
not the discrimination of the method, because a method can reach a high
concordance index and still have a large selection regret when its loss
cannot tell which of its own checkpoints ranks best. A loss whose
selection regret is small and stable across datasets is more useful than
one whose regret is small only on average, because a practitioner cannot
know in advance which dataset will produce a large regret. We report the
selection regret alongside the loss-metric rank correlation because it
is directly meaningful and is robust to the plateau noise that can
corrupt a raw rank correlation. The loss-metric rank correlation is the
primary quantity by which we compare losses, and it is defined here so
that the comparison in \hyperlink{sec:5}{Section 5} is unambiguous.

\hypertarget{sec:4}{}\subsection{4. Experimental setup}\label{experimental-setup}

\hypertarget{sec:4.1}{}\subsubsection{4.1 Datasets}\label{datasets}

We evaluate across four modalities and eighteen datasets, with two
independent cohorts for each imaging modality so that no conclusion
rests on a single dataset, and with a wide range of censoring so that
robustness to the fraction of observed events can be assessed. The
tabular modality comprises ten public clinical and genomic datasets:
METABRIC \cite{curtis2012}, SUPPORT \cite{knaus1995}, the Rotterdam and
German Breast Cancer Study Group cohorts \cite{schumacher1994}, FLCHAIN
\cite{dispenzieri2012}, WHAS500 \cite{hosmer2008}, the primary biliary
cholangitis cohort \cite{therneau2000}, the NCCTG lung cohort
\cite{loprinzi1994}, a Surveillance, Epidemiology, and End Results
cohort \cite{seer}, and a TCGA glioblastoma cohort \cite{weinstein2013}.
The computed tomography and positron emission tomography modality
comprises the HECKTOR 2026 head-and-neck cohort, with recurrence-free
survival as the endpoint, and the Colorectal-Liver-Metastases cohort
\cite{simpson2024}, obtained from The Cancer Imaging Archive
\cite{clark2013}, with overall survival as the endpoint. The magnetic
resonance imaging modality comprises the UCSF-PDGM and UPENN-GBM glioma
cohorts, both with overall survival. The pathology modality comprises
four cohorts from The Cancer Genome Atlas \cite{weinstein2013}, namely
breast invasive carcinoma, the combined glioma cohort, lung
adenocarcinoma, and uterine corpus endometrial carcinoma, each
represented by slide-level features from the TITAN foundation model.
\hyperlink{tab:1}{Table 1} summarizes the cohorts, their sizes, event rates, endpoints, and
inputs.

\hypertarget{tab:1}{}\textbf{Table~1. Summary of the eighteen datasets used in the study,
grouped by modality.}

{\def\LTcaptype{none} 
\begin{longtable}[]{@{}
  >{\raggedright\arraybackslash}p{(\linewidth - 10\tabcolsep) * \real{0.1667}}
  >{\raggedright\arraybackslash}p{(\linewidth - 10\tabcolsep) * \real{0.1667}}
  >{\raggedright\arraybackslash}p{(\linewidth - 10\tabcolsep) * \real{0.1667}}
  >{\raggedright\arraybackslash}p{(\linewidth - 10\tabcolsep) * \real{0.1667}}
  >{\raggedright\arraybackslash}p{(\linewidth - 10\tabcolsep) * \real{0.1667}}
  >{\raggedright\arraybackslash}p{(\linewidth - 10\tabcolsep) * \real{0.1667}}@{}}
\toprule\noalign{}
\begin{minipage}[b]{\linewidth}\raggedright
Modality
\end{minipage} & \begin{minipage}[b]{\linewidth}\raggedright
Dataset
\end{minipage} & \begin{minipage}[b]{\linewidth}\raggedright
Patients
\end{minipage} & \begin{minipage}[b]{\linewidth}\raggedright
Event rate
\end{minipage} & \begin{minipage}[b]{\linewidth}\raggedright
Endpoint
\end{minipage} & \begin{minipage}[b]{\linewidth}\raggedright
Input to the model
\end{minipage} \\
\midrule\noalign{}
\endhead
\bottomrule\noalign{}
\endlastfoot
Tabular & Ten public cohorts (METABRIC, SUPPORT, GBSG, ROTTERDAM,
FLCHAIN, WHAS500, SEER, PBC, lung, TCGA-GBM) & 228 to 9,105 & 0.15 to
0.82 & Overall survival & Clinical and genomic feature vectors \\
CT / PET-CT & HECKTOR 2026 (head and neck) & 727 & 0.20 &
Recurrence-free survival & CT, PET, and tumor mask, three channels \\
CT / PET-CT & Colorectal-Liver-Metastases & 197 & 0.54 & Overall
survival & CT and tumor mask, two channels \\
MRI & UCSF-PDGM (glioma) & 499 & 0.50 & Overall survival & Four MRI
sequences and tumor mask \\
MRI & UPENN-GBM (glioblastoma) & 611 & 0.96 & Overall survival & Four
MRI sequences and tumor mask \\
Pathology & TCGA breast (950), glioma (573), lung (444), endometrial
(480) & 2,447 total & 0.14 to 0.35 & Overall survival & TITAN
slide-level features, 768 dimensions \\
\end{longtable}
}

For the three-dimensional imaging cohorts, volumes are cropped to a
tumor-centered cube of 128 millimeters guided by the segmentation and
resampled to a grid of 64 by 64 by 64 voxels at an isotropic spacing of
2 millimeters, which preserves physical scale across patients. Magnetic
resonance sequences are z-scored within the brain region on a
per-sequence basis, computed tomography intensities are windowed to a
fixed soft-tissue Hounsfield range of -100 to 300 for the
Colorectal-Liver-Metastases cohort, and the positron emission tomography
and computed tomography inputs of HECKTOR are used at their
challenge-provided normalization. Clinical covariates are concatenated
to the image features where available.

\hypertarget{sec:4.2}{}\subsubsection{4.2 Model architectures}\label{model-architectures}

The purpose of the study is to develop and evaluate a loss function, not
to advance any particular architecture, so for each modality we adopt a
standard and deliberately unremarkable backbone and change only the
loss. \hyperlink{fig:2}{Figure 2} shows this shared design, in which a modality-specific
backbone maps the input to a scalar risk score and the loss is the only
component that varies between methods.

\begin{figure}
\hypertarget{fig:2}{}\centering
\pandocbounded{\includegraphics[keepaspectratio,alt={\hyperlink{fig:2}{Figure 2}. Overview of the study design. For each modality a standard backbone maps the input to a scalar risk score, and only the loss function differs between methods. The sigmoid concordance loss (SCL) is computed over comparable patient pairs within each minibatch.}]{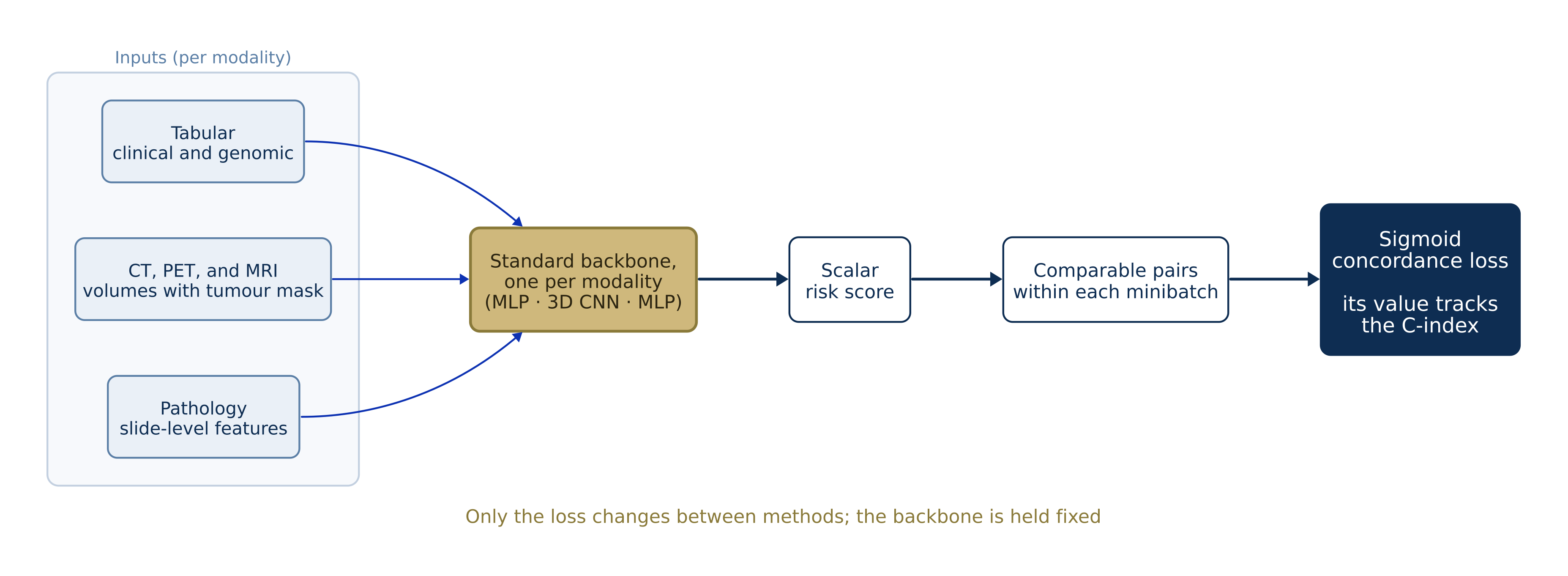}}
\caption{Figure~2. Overview of the study design. For each modality a
standard backbone maps the input to a scalar risk score, and only the
loss function differs between methods. The sigmoid concordance loss
(SCL) is computed over comparable patient pairs within each minibatch.}
\end{figure}

Tabular data use a multilayer perceptron with a scalar risk output. The
three-dimensional imaging cohorts use a compact three-dimensional
convolutional encoder that consumes the cropped multi-channel volume,
followed by fusion with clinical covariates and a scalar risk head. The
pathology cohorts use a small multilayer perceptron on the pre-pooled
slide-level features, which is the standard setting for pre-extracted
foundation-model features and which isolates the effect of the loss from
the choice of aggregator. Using a fixed backbone per modality ensures
that any difference between losses is attributable to the objective
rather than to the network, which is the comparison the paper is
designed to make.

\hypertarget{sec:4.3}{}\subsubsection{4.3 Baseline losses}\label{baseline-losses}

Against the SCL we compare the objectives that dominate the deep
survival literature, namely the Cox partial likelihood as used in
DeepSurv, the discrete-time negative log-likelihood of multi-task
logistic regression, the likelihood and ranking objective of DeepHit,
and the time-adaptive ranking objective of TripleSurv. On the tabular
modality we additionally include a random survival forest
\cite{ishwaran2008} as a non-neural reference. We also include the
hybrid loss of \hyperlink{sec:3.6}{Section 3.6} as a diagnostic ablation rather than as a
competitor. All neural methods share the same backbone within a
modality, and only the loss changes, so the comparison is controlled.

\hypertarget{sec:4.4}{}\subsubsection{4.4 Training protocol and evaluation
metrics}\label{training-protocol-and-evaluation-metrics}

Every dataset is evaluated with the same protocol. We use
event-stratified five-fold cross-validation, pool the out-of-fold
predictions to compute one estimate per metric, and repeat the whole
procedure over three random seeds, reporting the mean and standard
deviation across seeds. Each model is trained for one hundred epochs,
and the checkpoint at the best validation Harrell concordance index is
used for the test evaluation, which is the standard and favorable
protocol for every method. We report three discrimination metrics,
namely the Harrell concordance index, the truncated Uno concordance
index with the truncation set at the eightieth percentile of the event
times to stabilize the inverse-probability-of-censoring weighting in the
tail, and the time-dependent area under the curve. We report calibration
with the integrated Brier score \cite{graf1999}, computed from a Breslow
baseline survival function \cite{breslow1972} for the scalar-risk
methods and from the native predicted survival curves for the
discrete-time methods, where a lower integrated Brier score denotes
better calibration. Finally, we report the two coupling statistics
defined in \hyperlink{sec:3.7}{Section 3.7}, namely the loss-metric rank correlation and the
selection regret. Implementation uses PyTorch with the lifelines and
scikit-survival libraries for the survival metrics; the full training
configuration including batch sizes, the exact metric definitions, and
the baseline settings are given in Sections A, B, and C of the
Supplementary Material and released with the code. Per-dataset results
for all ten tabular cohorts and all four pathology cohorts, including
every metric reported below, are given in the Supplementary Material in
Tables S1 and S2.

\hypertarget{sec:5}{}\subsection{5. Results}\label{results}

\hypertarget{sec:5.1}{}\subsubsection{5.1 Discrimination and loss-metric
coupling}\label{discrimination-and-loss-metric-coupling}

Across all four modalities and all eighteen datasets we observe a
consistent pattern: the SCL is competitive on discrimination, being best
or within approximately one standard deviation of the best method on the
C-index, while being the only loss whose value stably tracks the C-index
during training. \hyperlink{tab:2}{Table 2} reports the full results across all four
modalities, giving for each method its discrimination, its within-cohort
C-index rank, its calibration, and its loss-metric coupling, and Figure
3 consolidates the loss-metric rank correlation across cohorts with
error bars.

\begin{figure}
\hypertarget{fig:3}{}\centering
\pandocbounded{\includegraphics[keepaspectratio,alt={\hyperlink{fig:3}{Figure 3}. Coupling between the loss and the concordance index across the six imaging and tabular cohorts, measured by the rank correlation between the negative validation loss and the validation concordance index. The SCL remains highly coupled on every cohort, while the likelihood losses couple weakly and with large variance. Error bars show the standard deviation across cross-validation repetitions.}]{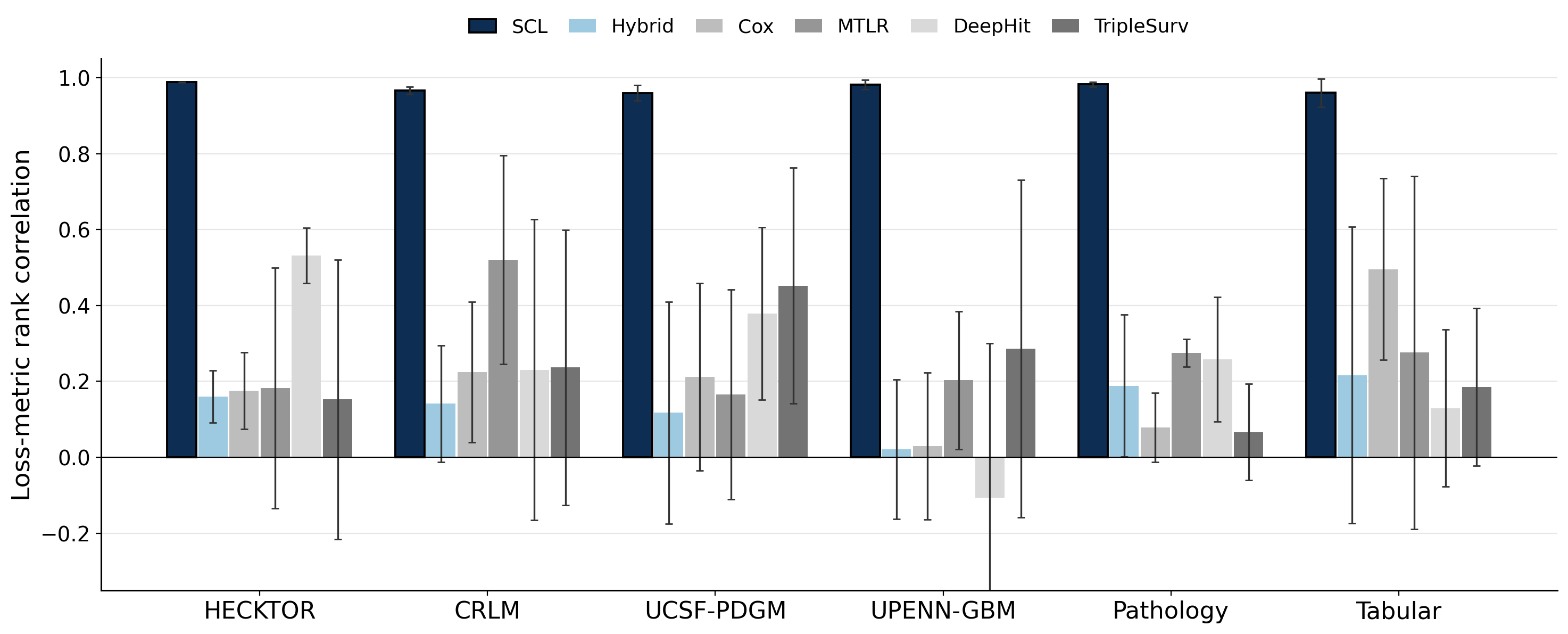}}
\caption{Figure~3. Coupling between the loss and the concordance index
across the six imaging and tabular cohorts, measured by the rank
correlation between the negative validation loss and the validation
concordance index. The SCL remains highly coupled on every cohort, while
the likelihood losses couple weakly and with large variance. Error bars
show the standard deviation across cross-validation repetitions.}
\end{figure}

On the tabular modality the SCL attains the best mean C-index rank
across the ten datasets, 2.60 out of 7, together with the highest
loss-metric rank correlation, 0.960. On the two computed tomography
cohorts it ties the best method on HECKTOR, 0.626 against 0.629 for
multi-task logistic regression, and is the best method on
Colorectal-Liver-Metastases, 0.607 against at most 0.586 for the
baselines, while its loss-metric rank correlation is 0.99 and 0.97
against at most 0.53 for the likelihood losses. On the two magnetic
resonance cohorts it is the best method on UPENN-GBM, 0.634, and is
within noise of the best on UCSF-PDGM, 0.691 against 0.708, while
coupling at 0.98 and 0.96. On the four pathology cohorts it ties the
field-standard Cox loss, 0.671 against 0.679, and attains the
second-best mean C-index rank across those cohorts behind Cox, with a
loss-metric rank correlation of 0.98 against at most 0.27 for the
baselines. In every cohort across the four modalities, whose event rates
span 12 to 96 per cent, the loss-metric rank correlation of the SCL is
between 0.96 and 0.99, whereas every likelihood-based baseline,
including the method that attains the best C-index on a given cohort,
has a low and high-variance coupling, which means that its loss is an
unreliable proxy for the C-index.

\hypertarget{tab:2}{}\textbf{Table~2. Master results across all four modalities under
five-fold cross-validation with three seeds.} SCL denotes our sigmoid
concordance loss and Hybrid its diagnostic variant with a likelihood
anchor (\hyperlink{sec:3.6}{Section 3.6}). HECKTOR and CRLM are CT/PET cohorts, UCSF-PDGM and
UPENN-GBM are MRI cohorts, and the Pathology and Tabular rows are means
over their four and ten datasets respectively. Methods are ordered by
Harrell concordance index, reported as the mean across seeds;
per-dataset standard deviations are given in the Supplementary Material.
The C-index rank is the within-cohort rank, or the mean within-dataset
rank for the pooled tabular and pathology blocks. A lower integrated
Brier score is better. The loss-metric rank correlation is the Spearman
correlation between the negative validation loss and the validation
concordance index over the later half of training; n/a marks a metric
undefined for the random survival forest.

{\def\LTcaptype{none} 
\begin{longtable}[]{@{}
  >{\raggedright\arraybackslash}p{(\linewidth - 14\tabcolsep) * \real{0.1250}}
  >{\raggedright\arraybackslash}p{(\linewidth - 14\tabcolsep) * \real{0.1250}}
  >{\raggedright\arraybackslash}p{(\linewidth - 14\tabcolsep) * \real{0.1250}}
  >{\raggedright\arraybackslash}p{(\linewidth - 14\tabcolsep) * \real{0.1250}}
  >{\raggedright\arraybackslash}p{(\linewidth - 14\tabcolsep) * \real{0.1250}}
  >{\raggedright\arraybackslash}p{(\linewidth - 14\tabcolsep) * \real{0.1250}}
  >{\raggedright\arraybackslash}p{(\linewidth - 14\tabcolsep) * \real{0.1250}}
  >{\raggedright\arraybackslash}p{(\linewidth - 14\tabcolsep) * \real{0.1250}}@{}}
\toprule\noalign{}
\begin{minipage}[b]{\linewidth}\raggedright
Cohort
\end{minipage} & \begin{minipage}[b]{\linewidth}\raggedright
Method
\end{minipage} & \begin{minipage}[b]{\linewidth}\raggedright
Harrell C-index
\end{minipage} & \begin{minipage}[b]{\linewidth}\raggedright
C-index rank
\end{minipage} & \begin{minipage}[b]{\linewidth}\raggedright
Truncated Uno
\end{minipage} & \begin{minipage}[b]{\linewidth}\raggedright
Time-dependent AUC
\end{minipage} & \begin{minipage}[b]{\linewidth}\raggedright
Integrated Brier score
\end{minipage} & \begin{minipage}[b]{\linewidth}\raggedright
Loss-metric rank correlation
\end{minipage} \\
\midrule\noalign{}
\endhead
\bottomrule\noalign{}
\endlastfoot
HECKTOR & MTLR & 0.629 & 1 & 0.635 & 0.655 & 0.113 & 0.18 \\
& \textbf{SCL} & 0.626 & 2 & 0.627 & 0.640 & 0.114 & 0.99 \\
& Hybrid & 0.617 & 3 & 0.619 & 0.623 & 0.132 & 0.16 \\
& Cox & 0.611 & 4 & 0.616 & 0.615 & 0.114 & 0.17 \\
& TripleSurv & 0.577 & 5 & 0.579 & 0.591 & 0.122 & 0.15 \\
& DeepHit & 0.561 & 6 & 0.560 & 0.578 & 0.153 & 0.53 \\
CRLM & \textbf{SCL} & 0.607 & 1 & 0.616 & 0.651 & 0.189 & 0.97 \\
& Cox & 0.586 & 2 & 0.593 & 0.624 & 0.191 & 0.23 \\
& MTLR & 0.577 & 3 & 0.577 & 0.612 & 0.197 & 0.52 \\
& Hybrid & 0.576 & 4 & 0.577 & 0.608 & 0.230 & 0.14 \\
& DeepHit & 0.567 & 5 & 0.566 & 0.586 & 0.217 & 0.23 \\
& TripleSurv & 0.561 & 6 & 0.566 & 0.579 & 0.191 & 0.24 \\
UCSF-PDGM & Hybrid & 0.708 & 1 & 0.697 & 0.758 & 0.168 & 0.12 \\
& MTLR & 0.707 & 2 & 0.696 & 0.758 & 0.167 & 0.17 \\
& \textbf{SCL} & 0.691 & 3 & 0.679 & 0.743 & 0.160 & 0.96 \\
& Cox & 0.683 & 4 & 0.673 & 0.733 & 0.160 & 0.21 \\
& DeepHit & 0.681 & 5 & 0.671 & 0.728 & 0.176 & 0.38 \\
& TripleSurv & 0.665 & 6 & 0.658 & 0.708 & 0.160 & 0.45 \\
UPENN-GBM & \textbf{SCL} & 0.634 & 1 & 0.637 & 0.690 & 0.171 & 0.98 \\
& TripleSurv & 0.631 & 2 & 0.634 & 0.684 & 0.166 & 0.29 \\
& DeepHit & 0.626 & 3 & 0.630 & 0.679 & 0.171 & -0.11 \\
& Hybrid & 0.622 & 4 & 0.625 & 0.674 & 0.175 & 0.02 \\
& MTLR & 0.620 & 5 & 0.623 & 0.669 & 0.170 & 0.20 \\
& Cox & 0.610 & 6 & 0.613 & 0.658 & 0.170 & 0.03 \\
Pathology & Cox & 0.679 & 2.00 & 0.669 & 0.688 & 0.140 & 0.08 \\
& \textbf{SCL} & 0.671 & 2.50 & 0.663 & 0.683 & 0.144 & 0.98 \\
& Hybrid & 0.675 & 2.75 & 0.662 & 0.683 & 0.193 & 0.19 \\
& MTLR & 0.650 & 4.25 & 0.649 & 0.662 & 0.146 & 0.27 \\
& TripleSurv & 0.624 & 4.50 & 0.620 & 0.638 & 0.143 & 0.07 \\
& DeepHit & 0.642 & 5.00 & 0.636 & 0.644 & 0.163 & 0.26 \\
Tabular & \textbf{SCL} & 0.785 & 2.60 & 0.786 & 0.821 & 0.131 & 0.96 \\
& RSF & 0.786 & 2.80 & 0.786 & 0.824 & n/a & n/a \\
& Hybrid & 0.780 & 3.40 & 0.782 & 0.817 & 0.274 & 0.22 \\
& MTLR & 0.781 & 3.50 & 0.782 & 0.819 & 0.120 & 0.28 \\
& Cox & 0.781 & 4.10 & 0.783 & 0.816 & 0.130 & 0.50 \\
& DeepHit & 0.760 & 5.00 & 0.762 & 0.794 & 0.160 & 0.13 \\
& TripleSurv & 0.693 & 6.60 & 0.696 & 0.729 & 0.144 & 0.19 \\
\end{longtable}
}

On the UCSF-PDGM cohort the hybrid and multi-task logistic regression
edge the SCL on the C-index by approximately 0.017, which is within one
standard deviation, and on the pathology cohorts the Cox loss edges it
by approximately 0.008, again within noise. These two cases are the
basis for the deliberate framing that the SCL matches rather than beats
the state-of-the-art (SOTA) on discrimination. What separates it from
the SOTA in every cohort is the coupling: its value tracks the C-index,
and the likelihood losses' values do not. \hyperlink{fig:4}{Figure 4} shows representative
validation trajectories on METABRIC in which the value of the SCL moves
together with the validation C-index, whereas the Cox likelihood loss
drifts upward even as the C-index improves, so its value does not track
the metric.

\begin{figure}
\hypertarget{fig:4}{}\centering
\pandocbounded{\includegraphics[keepaspectratio,alt={\hyperlink{fig:4}{Figure 4}. Training and validation trajectories for six losses on the METABRIC cohort (1904 patients, 801 events). Each panel shows the data-fit training loss (dashed), the validation loss (solid), and the validation concordance index (gold, right axis) over epochs; the likelihood losses are shown without their weight penalty. The value above each panel is the Spearman rank correlation between the negative validation loss and the validation concordance index over the second half of training. The gold star and navy circle mark the checkpoints selected by the best validation concordance index and by the minimum validation loss. Only for the SCL do the two markers coincide; for the other losses the validation loss drifts while the concordance index plateaus. Remaining datasets appear in Figures S3 to S11.}]{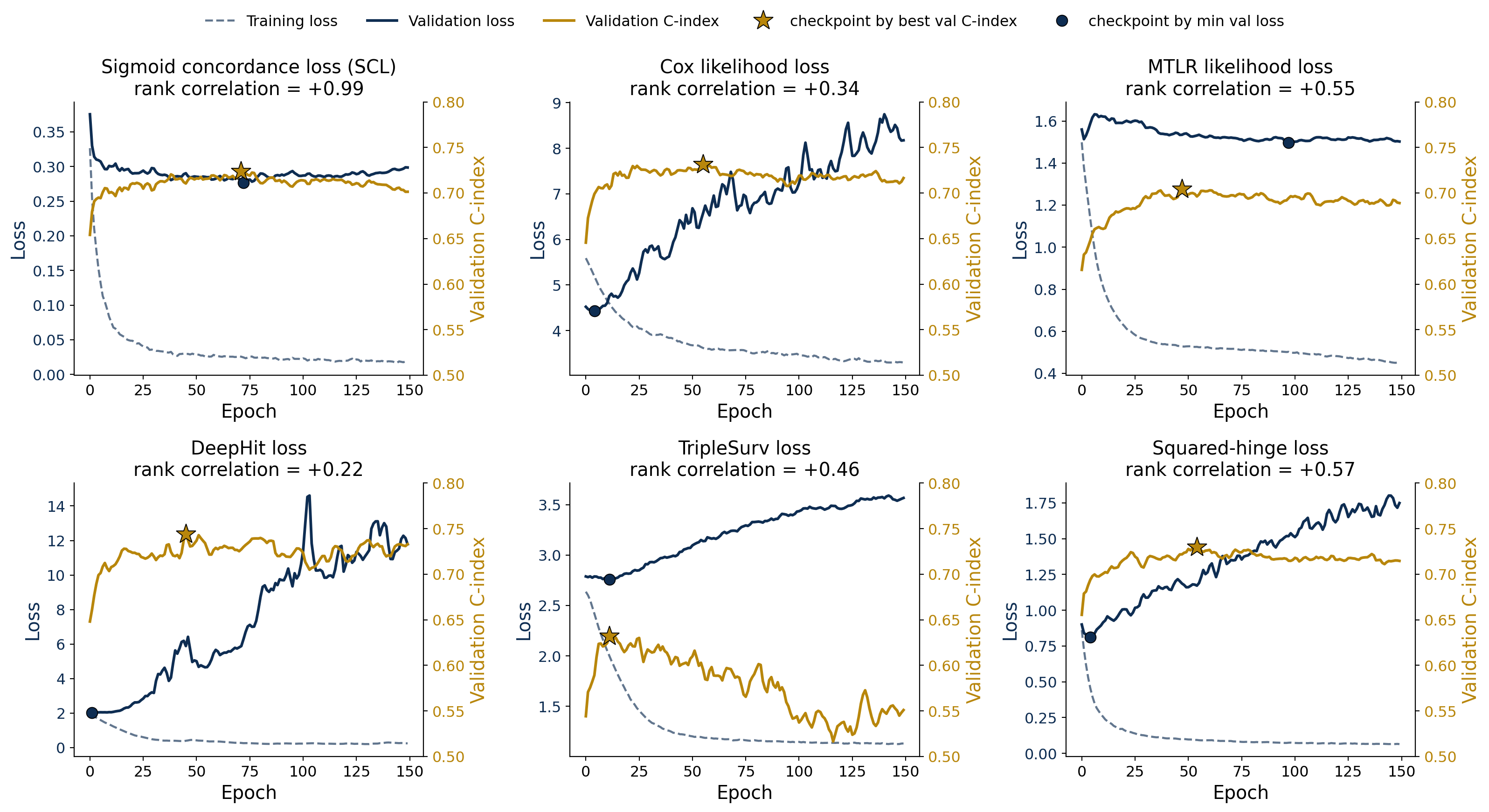}}
\caption{Figure~4. Training and validation trajectories for six losses
on the METABRIC cohort (1904 patients, 801 events). Each panel shows the
data-fit training loss (dashed), the validation loss (solid), and the
validation concordance index (gold, right axis) over epochs; the
likelihood losses are shown without their weight penalty. The value
above each panel is the Spearman rank correlation between the negative
validation loss and the validation concordance index over the second
half of training. The gold star and navy circle mark the checkpoints
selected by the best validation concordance index and by the minimum
validation loss. Only for the SCL do the two markers coincide; for the
other losses the validation loss drifts while the concordance index
plateaus. Remaining datasets appear in Figures S3 to S11.}
\end{figure}

The same trajectories for the remaining datasets across all three
imaging modalities and the tabular cohorts are given in Figures S3 to
S11 of the Supplementary Material, where the pattern is consistent.

\hypertarget{sec:5.2}{}\subsubsection{5.2 Calibration}\label{calibration}

Discrimination and coupling do not come at the expense of calibration.
The integrated Brier score, reported in \hyperlink{tab:2}{Table 2} alongside the
discrimination metrics, shows that the SCL, evaluated through a Breslow
baseline that turns its risk score into a survival function, is
comparable with the likelihood losses that model the survival
distribution directly. On the ten tabular datasets its mean integrated
Brier score of 0.131 is essentially tied with the Cox loss at 0.130 and
is close to the best method, multi-task logistic regression at 0.120. On
the imaging cohorts it is within a few thousandths of the
best-calibrated method, for example 0.114 on HECKTOR against 0.113 for
multi-task logistic regression, and on the four pathology cohorts it
averages 0.144 against 0.140 for Cox. A pure ranking loss therefore does
not sacrifice calibration once its scores are combined with a
nonparametric baseline, which addresses the natural concern that
optimizing order alone might yield poorly calibrated survival estimates.
The hybrid variant is the notable exception, with the worst calibration
on the tabular and pathology cohorts, 0.274 and 0.193, which is further
evidence that the hybrid is a diagnostic rather than a competitor, since
adding the likelihood anchor to the ranking loss both degrades coupling,
as shown in \hyperlink{sec:5.4}{Section 5.4}, and destabilizes calibration.

\hypertarget{sec:5.3}{}\subsubsection{5.3 Model selection and risk
stratification}\label{model-selection-and-risk-stratification}

Value-monotonicity has a direct practical payoff in model selection. A
method produces one checkpoint per epoch, and the checkpoint that is
deployed must be chosen on a validation signal, because the test
outcomes are held out. Choosing by the validation loss rather than by
the validation concordance index incurs the selection regret defined in
\hyperlink{sec:3.7}{Section 3.7}. \hyperlink{fig:5}{Figure 5} shows the distribution of this regret across all
eighteen datasets and four modalities, with one point per dataset. For
the SCL the regret never exceeds 0.011 of concordance index on any
dataset, with a standard deviation of 0.004, so its loss can be trusted
to select the checkpoint on every dataset we tried. The likelihood and
margin losses have a small average regret but are unpredictable: their
regret exceeds 0.02 on between one and eight of the eighteen datasets
and reaches 0.08 for DeepHit and 0.11 for TripleSurv in the worst case,
with no way to know in advance which dataset will be affected. Selecting
the model on the freely available loss is therefore safe for the SCL and
risky for the others, which matters most in the expensive end-to-end
regime where recomputing the concordance index at every epoch is the
dominant cost.

\begin{figure}
\hypertarget{fig:5}{}\centering
\pandocbounded{\includegraphics[keepaspectratio,alt={\hyperlink{fig:5}{Figure 5}. Distribution of the selection regret across all eighteen datasets and four modalities, with one point per dataset colored by modality. The selection regret is the test concordance index lost by choosing the deployed checkpoint on the validation loss instead of on the validation concordance index, and the shaded band marks a regret below 0.02. The SCL keeps the regret within 0.011 on every dataset, whereas the likelihood and margin losses have a small average regret but are unpredictable, exceeding 0.02 on many datasets and reaching about 0.08 for DeepHit and 0.11 for TripleSurv.}]{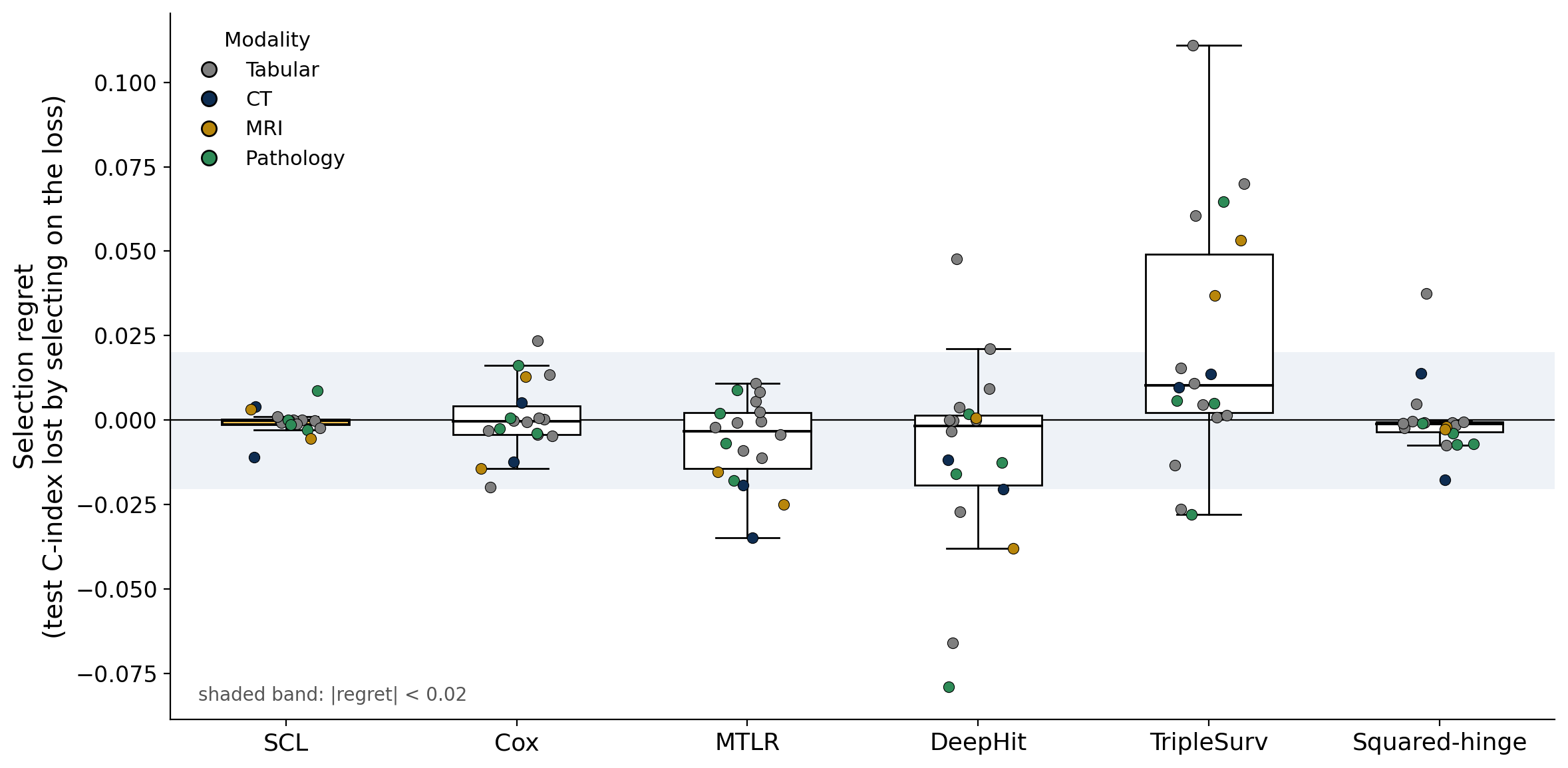}}
\caption{Figure~5. Distribution of the selection regret across all
eighteen datasets and four modalities, with one point per dataset
colored by modality. The selection regret is the test concordance index
lost by choosing the deployed checkpoint on the validation loss instead
of on the validation concordance index, and the shaded band marks a
regret below 0.02. The SCL keeps the regret within 0.011 on every
dataset, whereas the likelihood and margin losses have a small average
regret but are unpredictable, exceeding 0.02 on many datasets and
reaching about 0.08 for DeepHit and 0.11 for TripleSurv.}
\end{figure}

\hyperlink{tab:S4}{Table S4} in the Supplementary Material reports, for every cohort, the
pooled test concordance index obtained under each of the two selection
rules.

The risk scores learned under the SCL are also clinically meaningful.
Splitting patients at the median out-of-fold predicted risk stratifies
survival significantly on all four imaging cohorts, with hazard ratios
between the high-risk and low-risk groups of 2.86 on HECKTOR for
recurrence-free survival, 2.82 on UCSF-PDGM, 1.69 on UPENN-GBM, and 1.87
on Colorectal-Liver-Metastases for overall survival, all significant by
the log-rank test and with clearly separated Kaplan-Meier curves shown
in \hyperlink{fig:S1}{Figure S1} of the Supplementary Material. The loss therefore produces
risk strata that separate patient outcomes, which is the clinical use to
which a survival model is ultimately put.

\hypertarget{sec:5.4}{}\subsubsection{5.4 Ablation studies}\label{ablation-studies}

The choice of surrogate matters, and the ablations confirm the mechanism
proposed in the theory. Replacing the sigmoid with the squared-hinge, or
margin-based, surrogate preserves the C-index but collapses the
coupling, with the loss-metric rank correlation falling from
approximately 0.98 to approximately 0.4. This is exactly the prediction
of \hyperlink{sec:3.5}{Section 3.5}, because the margin term makes the value of the loss vary
within a class of scores that share an ordering. Adding a likelihood
anchor to the SCL, which produces the hybrid variant, reintroduces the
decoupling: the loss-metric rank correlation drops from approximately
0.98 to between 0.12 and 0.30, and there is no accompanying gain in the
C-index, which is the controlled confirmation that a proper-scoring-rule
term restores the order-preserving gradient direction. \hyperlink{fig:6}{Figure 6} sweeps
the anchor weight and shows that the coupling collapses as soon as any
positive anchor weight is added, falling from approximately 0.98 at zero
anchor to between 0.05 and 0.55 at small positive weights, while the
test C-index is essentially flat in the anchor weight. The dependence is
not monotone: the coupling is worst at intermediate anchor weights,
where the two terms most contend, and recovers only partially at large
weights as the objective approaches a pure likelihood, which itself has
moderate coupling.

\begin{figure}
\hypertarget{fig:6}{}\centering
\pandocbounded{\includegraphics[keepaspectratio,alt={\hyperlink{fig:6}{Figure 6}. Effect of the likelihood anchor weight in the hybrid loss. The loss-metric coupling collapses as soon as any positive anchor weight is added, while the test concordance index is essentially unchanged.}]{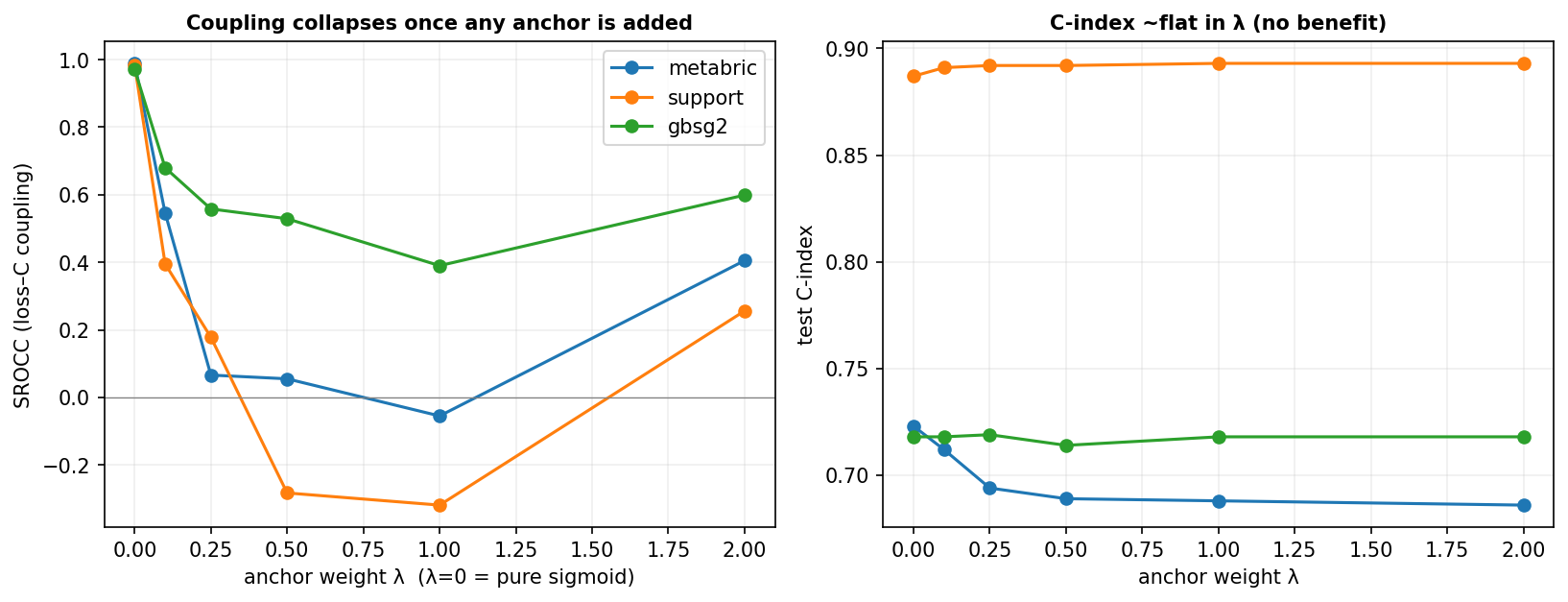}}
\caption{Figure~6. Effect of the likelihood anchor weight in the hybrid
loss. The loss-metric coupling collapses as soon as any positive anchor
weight is added, while the test concordance index is essentially
unchanged.}
\end{figure}

Finally, under increasing training censoring the ranking loss degrades
least among the losses considered, as shown in \hyperlink{fig:S2}{Figure S2} of the
Supplementary Material.

\hypertarget{sec:6}{}\subsection{6. Discussion}\label{discussion}

The results support a reframing of how a survival loss should be judged.
In the survival literature a loss is evaluated by the C-index that it
can reach, and by that measure the SCL and the standard likelihood
losses are comparable, with the SCL best or within roughly one standard
deviation of the best method across the four modalities, and with a
likelihood loss edging it by approximately 0.01 to 0.02 on two of them.
The C-index that a loss can reach and the C-index that its value tracks
during training are, however, different properties, and the second is
the one that governs whether the loss can be trusted before convergence.
On that second property the SCL and the likelihood losses are not
comparable at all. The SCL couples to the C-index at a rank correlation
of approximately 0.96 to 0.99 with tight variance, and it selects models
at close to zero regret, while every likelihood loss couples weakly and
with high variance and incurs nonzero selection regret. The contribution
of the paper is therefore not a new state of the art on discrimination
but a loss whose value can be used as a proxy for the model evaluation
metric during training.

The training trajectories in \hyperlink{fig:4}{Figure 4} make this concrete and connect it
to checkpoint selection. The training loss falls for every method, so
optimization proceeds as intended, but the validation loss behaves very
differently across losses. For the Cox, DeepHit, TripleSurv, and
squared-hinge losses the validation loss rises above the training loss
as training continues, which is ordinary overfitting, because the model
keeps sharpening the score magnitudes, margins, and predicted
distributions that the concordance index does not depend on. For the
multi-task logistic regression loss the validation loss instead keeps
decreasing. In both patterns the validation concordance index has
already reached a plateau, so the loss and the metric have come apart,
and the overfitting that the validation loss reports is invisible to the
concordance index. A practitioner who watches only the loss therefore
reads continued change, either progress or deterioration, where the
ranking is in fact stable. Such a practitioner either stops at the wrong
epoch or keeps the checkpoint whose loss looks best, which is not the
checkpoint whose ranking is best. The SCL removes this failure mode,
because its value equals one minus the concordance index up to the
temperature term, so its validation trajectory rises and falls with the
metric and the checkpoint of minimum validation loss is essentially the
checkpoint of maximum validation concordance index. On the cohort shown,
selecting the model by validation loss alone costs no test concordance
index for the SCL, whereas it costs between one and six points of
concordance index for the likelihood and margin losses, so a model
selected on the SCL is competitive with, and here better than, a model
selected on a decoupled loss that is given full access to the
concordance index. This is the practical content of value-monotonicity.
The loss can serve as the selection signal itself, which is exactly what
the expensive end-to-end regime forces the practitioner to rely on.

A reasonable objection to the practical value of this result is that the
measured selection regret is modest, often between 0.01 and 0.03 of
C-index and at most approximately 0.05, so that the practical cost of a
decoupled loss appears small. Three considerations answer this. The
first is that the value of a coupled loss is reliability rather than
magnitude. A loss that provably tracks the C-index removes the need to
compute the metric on a full validation set at every epoch, which in
end-to-end imaging is the dominant cost of evaluation, so a coupled loss
is a cheap and trustworthy substitute, whereas a decoupled loss forces
the expensive evaluation that one was trying to avoid. The second is
that the regret concentrates where it is least affordable, namely on the
small and heavily censored cohorts and late in training, which is
precisely the medical-imaging regime, where a swing of several points of
C-index can reverse the apparent ranking of a model against its
baselines. The third is that our protocol flatters the baselines on this
axis, because five-fold cross-validation with many logged checkpoints
allows us to recover the genuinely best-C-index model for every method,
which hides the cost of an untrustworthy loss; a practitioner with a
single expensive end-to-end run and sparse evaluation has no such safety
net, so the deployed regret is an underestimate of the practical cost.
Value-monotonicity is, in effect, insurance whose premium is zero,
because the SCL ties on the C-index, and whose payout is realized in the
setting that this paper targets.

The hybrid variant is best understood as a controlled test of the theory
rather than as a method. Adding a strictly proper likelihood anchor to
the SCL reintroduces the order-preserving gradient direction that
\hyperlink{sec:3.5}{Section 3.5} identifies, so the theory predicts that the anchor must
degrade coupling, and the anchor-weight sweep confirms this sharply,
with coupling collapsing as soon as any positive anchor weight is added
and with no compensating gain in the C-index. The prediction is not
merely qualitative but has the non-monotone shape that the mechanism
implies, with the worst coupling at intermediate anchor weights. This
gives the theory predictive rather than merely descriptive content, and
it is the reason the hybrid is reported at all. The practical conclusion
is that no positive anchor weight buys discrimination while preserving
coupling, so only the pure SCL is value-monotone.

The coupling statistic itself deserves a word of caution. A raw Spearman
correlation between the loss trajectory and the metric trajectory is
scale-free but not aware of magnitude, so on a converged plateau it
measures noise and can take spuriously negative values. For this reason
we lead with the selection regret, which is robust and directly
meaningful, and we report the rank correlation as a magnitude of
coupling restricted to the informative regime. Strongly negative
correlations should not be read as systematic inversion, and the
defensible statement supported by the data is that the SCL stays
informatively coupled while the likelihood losses become uninformative.

The study has several limitations. The general form of the decoupling
theorem is proved for arbitrary order-preserving directions through a
group-orbit argument, and the always-available rescaling case is made
fully explicit, but a finite-sample coupling bound for the
value-monotonicity theorem, which would follow from the concentration of
the loss as a U-statistic over comparable pairs, is sketched rather than
written in full. The experiments vary the loss on a fixed backbone per
modality in order to isolate its effect, so we do not co-tune the
architecture and the loss, and the pathology experiments use pre-pooled
features rather than end-to-end multiple-instance learning, so the
extension to end-to-end gigapixel training remains future work. The
temperature trades smoothness against the residual error in the
value-tracking approximation, and we fix it rather than studying its
optimal schedule. One weak baseline produced a degenerate, constant
validation-loss trajectory on the magnetic resonance cohorts, which
leaves its coupling statistic undefined there. Finally, small event
counts inflate the variance of the Uno concordance index, which we
mitigate by truncation and by pooled cross-validation, although the
imaging cohorts remain modest in size.

The mechanism is not specific to survival analysis. It applies whenever
a rank-based evaluation metric such as the area under the curve, the
concordance index, or the normalized discounted cumulative gain is
optimized through a proper scoring rule, because it is the
rank-invariance of the metric together with the non-invariance of the
proper score that forces the decoupling, and a value-monotone surrogate
is the general remedy. Survival prediction is a clean and clinically
consequential instance of this broader phenomenon.

\hypertarget{sec:7}{}\subsection{7. Conclusion}\label{conclusion}

We have shown that the mismatch between concordance-based evaluation and
likelihood-based training is structural rather than incidental. Every
strictly proper survival likelihood admits an explicit direction along
which the loss decreases while the C-index is exactly unchanged, so the
value of a likelihood loss decouples from the metric during training,
most severely once the ranking has converged and the loss is trusted
most. The SCL is value-monotone in the C-index by construction, its
value equal to one minus the C-index up to a temperature term, and
across tabular, computed tomography, magnetic resonance, and pathology
data it matches the discrimination of the standard likelihood losses
while being the only loss whose value stably tracks the C-index and
supports near-zero-regret model selection, at comparable calibration and
with significant clinical risk stratification. This is what makes the
loss a trustworthy guide for monitoring, early stopping, and model
selection when an encoder is trained end to end on scarce clinical data
and the concordance index cannot be recomputed at every epoch. We
release the loss, the preprocessing pipelines for all four modalities,
and the full evaluation protocol as open-source code at
\url{https://github.com/Meixu-Chen/sigmoid-concordance-loss}.

\subsection{Acknowledgments}\label{acknowledgments}

This work was partially supported by the National Institutes of Health
(NIH; R01 CA251792) and a 2024 American Association of Physicists in
Medicine (AAPM) Research Seed Funding Grant.

\subsection{Appendix A. Theoretical results and
proofs}\label{appendix-a.-theoretical-results-and-proofs}

This appendix states and proves the two claims behind the method. The
first is that the SCL is value-monotone in the concordance index, so its
value tracks the metric across training and not only at its minimizer.
The second is that strictly proper survival likelihood losses are not
value-monotone and provably decouple from the concordance index, most
severely late in training. Both follow from a single structural fact,
namely that the concordance index is invariant to strictly increasing
reparameterizations of the risk score.

\subsubsection{A.1 Setup}\label{a.1-setup}

The data are triples \((X, T, \delta)\) with covariates
\(X \in \mathcal{X}\), event or censoring time \(T > 0\), and event
indicator \(\delta \in \{0,1\}\). A model is a risk score
\(f : \mathcal{X} \to \mathbb{R}\), where a higher score denotes higher
risk and shorter survival, and we write \(f_i = f(X_i)\). A pair
\((i,j)\) is comparable if \(T_i < T_j\) and \(\delta_i = 1\), so that
the earlier subject is known to have failed first, and we write
\(\mathcal{P}\) for the distribution of comparable pairs. The population
concordance index is \[
C(f) = \Pr\big( f_i > f_j \mid (i,j)\ \text{comparable} \big) = \mathbb{E}_{(i,j) \sim \mathcal{P}} \big[ \mathbf{1}\{ f_i > f_j \} \big],
\] with ties split as one half. The Harrell and Uno versions differ only
in how the inverse-probability-of-censoring weights reweight
\(\mathcal{P}\), which does not affect any argument below.

\textbf{Assumption A1 (non-atomic gap).} The score gap \(Z = f_i - f_j\)
under \(\mathcal{P}\) has a density bounded by \(B < \infty\) in a
neighborhood of zero. This holds whenever \(f\) is non-degenerate and
\(X\) has a density, and it rules out a point mass exactly at ties. When
a class of scores \(\mathcal{F}\) is considered, as in Proposition 3, we
assume this bound holds uniformly over \(\mathcal{F}\).

\subsubsection{A.2 The structural fact}\label{a.2-the-structural-fact}

\textbf{Lemma 1 (rank-invariance of the concordance index).} For every
strictly increasing function \(\varphi : \mathbb{R} \to \mathbb{R}\),
\(C(\varphi \circ f) = C(f)\).

\emph{Proof.} Since \(\varphi\) is strictly increasing,
\(\varphi(f_i) > \varphi(f_j)\) if and only if \(f_i > f_j\), so the
event inside the expectation is unchanged. \(\square\)

Lemma 1 says that the concordance index factors through the ordering of
\(f\) alone, and that its level sets are the classes of scores that
share an ordering. A loss is faithful to the concordance index precisely
when, informally, its value does not move within such a class. Define
the rank-preserving tangent cone at a score \(f\) as the set of
directions that change \(f\) without changing \(C\) to first order,
\(R_f = \{ h : \tfrac{d}{dt} C(f + t h)\big|_{t=0} = 0 \}\).

\subsubsection{A.3 The SCL is
value-monotone}\label{a.3-the-scl-is-value-monotone}

For temperature \(\tau > 0\) the population SCL over comparable pairs is
\[
L_\tau(f) = \mathbb{E}_{(i,j) \sim \mathcal{P}} \big[ \sigma( -(f_i - f_j)/\tau ) \big], \qquad \sigma(u) = \frac{1}{1 + e^{-u}}.
\]

\textbf{Theorem 1 (value-monotonicity).} Under Assumption A1, \[
L_\tau(f) = \big( 1 - C(f) \big) + O(\tau) \quad \text{as } \tau \to 0,
\] uniformly, with a constant at most \(2B\). Consequently, for small
\(\tau\), the value of \(L_\tau\) is a strictly decreasing function of
\(C\) up to a term of order \(\tau\), so lower loss corresponds to
higher concordance index, and the value of the loss tracks the metric
rather than merely sharing its optimum.

\emph{Proof.} Let \(Z = f_i - f_j\) have density \(p\). Then
\(L_\tau(f) = \mathbb{E}[\sigma(-Z/\tau)]\) and
\(1 - C(f) = \mathbb{E}[\mathbf{1}\{Z < 0\}]\) up to the measure-zero
tie set under Assumption A1. Their difference is
\(\mathbb{E}[g_\tau(Z)]\) with
\(g_\tau(z) = \sigma(-z/\tau) - \mathbf{1}\{z < 0\}\), which satisfies
\(|g_\tau(z)| \le e^{-|z|/\tau}\), because for \(z > 0\) we have
\(\sigma(-z/\tau) \le e^{-z/\tau}\) and for \(z < 0\) we have
\(1 - \sigma(-z/\tau) = \sigma(z/\tau) \le e^{z/\tau}\). Hence
\(|L_\tau - (1 - C)| \le \int e^{-|z|/\tau} p(z)\, dz \le 2 B \tau + o(\tau)\)
by bounding the density near zero and the tails separately. \(\square\)

\textbf{Corollary 1 (Fisher consistency).}
\(\arg\min_f \lim_{\tau \to 0} L_\tau(f) = \arg\max_f C(f)\), so
minimizing the surrogate targets the concordance-index maximizer, and by
Theorem 1 it does so along a path whose value mirrors the metric.

\emph{Why the sigmoid and not the squared hinge.} The squared-hinge, or
margin-based, surrogate
\(L_{\mathrm{hinge}}(f) = \mathbb{E}[(m - (f_i - f_j))_+^2]\) is
consistent for the area under the curve \cite{gao2015} and therefore
shares Corollary 1, but it is not value-monotone. It depends on the
margins \(f_i - f_j\), so it varies within a class of scores that share
an ordering, and
\(L_{\mathrm{hinge}}(\varphi \circ f) \ne L_{\mathrm{hinge}}(f)\) for a
nonlinear increasing \(\varphi\). Its value is therefore not a function
of the concordance index, which explains the empirical gap between the
sigmoid and the squared hinge on coupling at comparable discrimination.

\subsubsection{A.4 Likelihood losses provably
decouple}\label{a.4-likelihood-losses-provably-decouple}

A survival likelihood loss, whether the Cox partial likelihood, a
discrete-time or logistic-hazard negative log-likelihood, or the
likelihood term of DeepHit, is a strictly proper scoring rule, so its
population minimizer is the true conditional law of the event given the
covariates, a single point in distribution space rather than a class of
orderings. Let \(G\) be the group of strictly increasing continuously
differentiable reparameterizations acting on scores by
\(\varphi \cdot f = \varphi \circ f\). By Lemma 1 the concordance index
is invariant under \(G\), so \(C(\varphi \cdot f) = C(f)\) for all
\(\varphi \in G\). Write \(G \cdot f\) for the orbit of \(f\), that is
its class of order-equivalent scores, and \(T_f(G \cdot f)\) for the
tangent space of that orbit at \(f\), which consists of the directions
that reshape values while preserving order.

\textbf{Theorem 2 (general decoupling).} Let \(L\) be
Gateaux-differentiable and not invariant under \(G\), which every
strictly proper survival likelihood satisfies, for example because
\(L(2f) \ne L(f)\) by Proposition 2 below. Then the orbit tangent space
is contained in the rank-preserving cone,
\(T_f(G \cdot f) \subseteq R_f = \ker dC(f)\), and at any \(f\) that is
not a critical point of \(L\) restricted to its orbit, the gradient
\(\nabla L(f)\) has a nonzero component in \(T_f(G \cdot f)\).
Equivalently, there is a descent direction of \(L\) along which the
concordance index is first-order stationary, so gradient flow on \(L\)
moves within the class of order-equivalent scores while the concordance
index does not change.

\emph{Proof.} For any differentiable curve
\(\alpha \mapsto \varphi_\alpha\) in \(G\) with \(\varphi_0\) the
identity, the map \(\alpha \mapsto C(\varphi_\alpha \circ f)\) is
constant by invariance, so differentiating at \(\alpha = 0\) gives
\(dC(f)[\xi] = 0\) for
\(\xi = \tfrac{d}{d\alpha}(\varphi_\alpha \circ f)\big|_{0} \in T_f(G \cdot f)\),
which shows \(T_f(G \cdot f) \subseteq \ker dC(f)\). Since \(L\) is not
invariant under \(G\), the map
\(\alpha \mapsto L(\varphi_\alpha \circ f)\) is non-constant for some
such curve, so
\(\tfrac{d}{d\alpha} L(\varphi_\alpha \circ f)\big|_{0} = dL(f)[\xi] \ne 0\)
unless \(f\) is a critical point of \(L\) restricted to its orbit. That
direction, or its sign flip, is a descent direction of \(L\) lying in
\(R_f\). \(\square\)

The always-available instance of this mechanism is rescaling, which we
make fully explicit and rigorous.

\textbf{Proposition 2 (rescaling-direction decoupling).} Let
\(L_{\mathrm{cox}}\) be the negative Cox partial log-likelihood on a
sample with event set \(D\) and risk sets \(R(i)\), and let \(f\) be any
risk score that is non-constant on some risk set. Define
\(\phi(\alpha) = L_{\mathrm{cox}}(\alpha f)\) for \(\alpha > 0\). Then
\(C(\alpha f) = C(f)\) for all \(\alpha > 0\) by Lemma 1, so the
direction \(h = f\) is exactly rank-preserving with \(dC(f)[f] = 0\);
and \(\phi\) is smooth and strictly convex, with \[
\phi''(\alpha) = \frac{1}{|D|} \sum_{i \in D} \mathrm{Var}_{p_\alpha^{(i)}}[f] > 0, \qquad p_\alpha^{(i)} = \mathrm{softmax}(\alpha f)\ \text{over}\ R(i),
\] so \(\phi'(1) \ne 0\) for all \(f\) outside a measure-zero set; when
\(f\) perfectly orders every risk set \(\phi\) is strictly decreasing
with no finite minimizer, and otherwise \(\phi\) has a unique interior
minimizer generically different from \(\alpha = 1\). Therefore the
rescaling axis is an explicit, always-available direction along which
the Cox likelihood strictly decreases while the concordance index is
exactly constant.

\emph{Proof.} The concordance identity is Lemma 1. Writing
\(\phi(\alpha) = -|D|^{-1} \sum_{i \in D} [\alpha f_i - \log \sum_{j \in R(i)} e^{\alpha f_j}]\)
gives
\(\phi'(\alpha) = -|D|^{-1} \sum_i [f_i - \mathbb{E}_{p_\alpha^{(i)}} f]\)
and
\(\phi''(\alpha) = |D|^{-1} \sum_i \mathrm{Var}_{p_\alpha^{(i)}} f \ge 0\),
strict whenever \(f\) is non-constant on some risk set. Strict convexity
yields at most one interior minimizer, and \(\phi'(1) = 0\) only on a
measure-zero set; if \(f\) perfectly orders every risk set then \(\phi\)
is strictly decreasing and \(\phi'(1) < 0\). The same computation holds
for any strictly proper discrete-time negative log-likelihood, whose
per-interval Bernoulli terms are likewise not scale invariant.
\(\square\)

Numerically, on an eight-hundred-sample synthetic Cox dataset, sweeping
the scale over a wide range leaves the concordance index with a range of
zero to five decimal places while the Cox negative log-likelihood ranges
from about 5.7 to about 14.2, with its minimum at an intermediate scale,
which is the phenomenon shown in \hyperlink{fig:2}{Figure 2} of the main text. Any
optimizer that adjusts the effective scale of the score, through the
weight norm, a temperature, or a normalization-layer scale, moves the
likelihood without moving the concordance index.

\textbf{Corollary 2 (uninformativeness late in training).} Along a
training trajectory, once the ranking has approximately converged so
that the rank-changing gradient component has vanished, the remaining
likelihood gradient lies predominantly in the rank-preserving subspace.
The likelihood therefore keeps decreasing while the concordance index
moves only through finite-sample noise, so the value of the loss becomes
uninformative about the metric. We state this as uninformativeness
rather than as systematic anti-correlation, because a strongly negative
rank correlation measured over such a converged window is a noise
artefact rather than a genuine inversion. The defensible statement
supported by the data is that the sigmoid surrogate stays informatively
coupled, with a rank correlation near the high nineties in hundredths
and tight variance, while every likelihood loss becomes uninformative,
with a low and high-variance rank correlation and nonzero selection
regret.

\subsubsection{A.5 A finite-sample value-monotonicity
bound}\label{a.5-a-finite-sample-value-monotonicity-bound}

Theorem 1 is a population statement. In practice we minimize, and
monitor, the empirical loss over the observed comparable pairs, so what
matters for the training-time argument is whether the empirical loss
value tracks the concordance index on a finite sample and uniformly over
the model class explored during optimization. We give the corresponding
finite-sample bound. Let \(\mathcal{F}\) be a class of risk scores with
\(\sup_{x} |f(x)| \le M\) for all \(f \in \mathcal{F}\), let
\(\sigma_\tau(u) = \sigma(-u/\tau)\), and let the empirical loss and
empirical concordance over the sample of \(n\) subjects be \[
\widehat{L}_\tau(f) = \frac{1}{|\widehat{\mathcal{P}}|} \sum_{(i,j) \in \widehat{\mathcal{P}}} \sigma_\tau\big(f_i - f_j\big), \qquad \widehat{C}(f) = \frac{1}{|\widehat{\mathcal{P}}|} \sum_{(i,j) \in \widehat{\mathcal{P}}} \mathbf{1}\{f_i > f_j\},
\] where \(\widehat{\mathcal{P}}\) is the set of observed comparable
pairs.

\textbf{Proposition 3 (finite-sample value-monotonicity).} Under
Assumption A1, with probability at least \(1 - \delta\) over the sample,
uniformly over \(f \in \mathcal{F}\), \[
\big| \widehat{L}_\tau(f) - (1 - C(f)) \big| \le 2 B \tau + \frac{C_0}{\tau}\, \mathcal{R}_n(\mathcal{F}) + \sqrt{\frac{\log(2/\delta)}{2 \lfloor n/2 \rfloor}},
\] where \(\mathcal{R}_n(\mathcal{F})\) is the Rademacher complexity
\cite{bartlett2002} of \(\mathcal{F}\) and \(C_0\) is a universal
constant. The empirical loss therefore estimates one minus the
concordance index uniformly over the class, with the error controlled by
the temperature through the approximation term \(2 B \tau\) and by the
capacity of \(\mathcal{F}\) and the sample size through the estimation
terms. The empirical loss trajectory tracks the concordance index up to
these terms, which is the finite-sample form of value-monotonicity.

\emph{Proof.} By the triangle inequality, \[
\big| \widehat{L}_\tau(f) - (1 - C(f)) \big| \le \underbrace{\big| \widehat{L}_\tau(f) - L_\tau(f) \big|}_{\text{estimation}} + \underbrace{\big| L_\tau(f) - (1 - C(f)) \big|}_{\text{approximation}}.
\] The approximation term is at most \(2 B \tau\) uniformly, by Theorem
1. For the estimation term, \(\widehat{L}_\tau(f)\) is a two-sample
U-statistic over comparable pairs with kernel
\(\sigma_\tau(f_i - f_j) \in [0,1]\); strictly it is a ratio of
U-statistics, since the comparable set and its cardinality
\(|\widehat{\mathcal{P}}|\) are random, and conditioning on the
comparable set reduces it to a U-statistic whose normalization
contributes a lower-order term that we suppress. By Hoeffding's
reduction \cite{hoeffding1963} any U-statistic of order two can be
written as an average over permutations of the data of averages of
\(\lfloor n/2 \rfloor\) independent and identically distributed terms.
Applying the bounded-differences inequality \cite{mcdiarmid1989} to each
such average of independent terms, followed by symmetrization over the
sup within \(\mathcal{F}\), gives \[
\sup_{f \in \mathcal{F}} \big| \widehat{L}_\tau(f) - L_\tau(f) \big| \le 2\, \mathcal{R}_n(\sigma_\tau \circ \Delta \mathcal{F}) + \sqrt{\frac{\log(2/\delta)}{2 \lfloor n/2 \rfloor}} \quad \text{with probability } 1 - \delta,
\] where
\(\Delta \mathcal{F} = \{ (x, x') \mapsto f(x) - f(x') : f \in \mathcal{F} \}\).
Since \(\sigma_\tau\) is \(\tfrac{1}{4\tau}\)-Lipschitz, Talagrand's
contraction inequality \cite{ledoux1991} gives
\(\mathcal{R}_n(\sigma_\tau \circ \Delta \mathcal{F}) \le \tfrac{1}{4\tau}\, \mathcal{R}_n(\Delta \mathcal{F})\),
and the pairwise-difference class satisfies
\(\mathcal{R}_n(\Delta \mathcal{F}) \le 2\, \mathcal{R}_n(\mathcal{F})\)
by subadditivity of the Rademacher complexity under the map
\(f(x) - f(x')\). Collecting these factors into a universal constant
\(C_0\) yields the stated estimation term, and combining with the
approximation term completes the proof. \(\square\)

The same U-statistic argument applied to the indicator kernel shows that
\(\widehat{C}(f)\) concentrates around \(C(f)\) at the rate
\(\sqrt{\log(1/\delta) / \lfloor n/2 \rfloor}\), so the empirical
concordance and the empirical loss track the same population quantity,
and their trajectories move together up to the temperature and
complexity terms above. The bound is stated with a universal constant
rather than optimized constants; sharper pairwise concentration is
available through the moment method for U-processes but is not needed
for the qualitative training-time conclusion.

\subsubsection{A.6 Remarks on scope}\label{a.6-remarks-on-scope}

Proposition 2 makes the rescaling case rigorous, which already
guarantees that decoupling is unavoidable for every model. The general
Theorem 2 extends this to arbitrary rank-preserving directions through
the orbit-tangent argument. A fully formal function-space statement that
a nonzero rank-preserving gradient component persists for arbitrary
directions, rather than only for rescaling, is a strengthening in the
same spirit, which we do not pursue here because the rescaling case
already establishes the core claim.

\bibliographystyle{unsrtnat}
\bibliography{refs}

\clearpage

\section{Supplementary Material}\label{supplementary-material}

\subsection{A value-monotone concordance loss for trustworthy deep
survival
prediction}\label{a-value-monotone-concordance-loss-for-trustworthy-deep-survival-prediction}

This supplement reports per-dataset results that are summarized in the
main text. All numbers come from the same five-fold cross-validation
protocol with three seeds, reported as mean plus or minus standard
deviation across seeds. The loss-metric rank correlation is the Spearman
correlation between the negative validation loss and the validation
concordance index over the later half of training. Lower integrated
Brier score denotes better calibration.

\subsection{A. Implementation details}\label{a.-implementation-details}

All methods share the training protocol below; only the loss function
differs. The goal of the study is to evaluate the loss, so each modality
uses a standard, deliberately unremarkable backbone.

\textbf{Cross-validation.} Event-stratified five-fold cross-validation;
out-of-fold predictions are pooled to compute one estimate per metric;
repeated over three random seeds and reported as mean plus or minus
standard deviation.

\textbf{Optimization.} Adam with learning rate 0.001 and weight decay
0.0001, one hundred epochs. The validation set is evaluated every two
epochs, and the checkpoint with the best validation Harrell concordance
index is used for testing.

\textbf{Loss.} The SCL uses a temperature of 0.1, which is fixed and not
tuned. The discrete-time methods (multi-task logistic regression and
DeepHit) bin the time axis at the quantiles of the event times, giving
roughly equal numbers of events per bin.

\textbf{Batch size} (the pairwise loss forms comparable pairs within
each batch, so larger is better where memory allows):

{\def\LTcaptype{none} 
\begin{longtable}[]{@{}
  >{\raggedright\arraybackslash}p{(\linewidth - 4\tabcolsep) * \real{0.3333}}
  >{\raggedright\arraybackslash}p{(\linewidth - 4\tabcolsep) * \real{0.3333}}
  >{\raggedright\arraybackslash}p{(\linewidth - 4\tabcolsep) * \real{0.3333}}@{}}
\toprule\noalign{}
\begin{minipage}[b]{\linewidth}\raggedright
Modality
\end{minipage} & \begin{minipage}[b]{\linewidth}\raggedright
Backbone
\end{minipage} & \begin{minipage}[b]{\linewidth}\raggedright
Batch size
\end{minipage} \\
\midrule\noalign{}
\endhead
\bottomrule\noalign{}
\endlastfoot
Tabular & Multilayer perceptron & min(256, max(32, n/4)), where n is the
training-fold size \\
CT and PET-CT & 3D convolutional encoder & 16 (limited by 3D volume
memory) \\
MRI & 3D convolutional encoder & 16 (limited by 3D volume memory) \\
Pathology & Multilayer perceptron on slide features & 64 \\
\end{longtable}
}

\textbf{Backbones.} Tabular: a multilayer perceptron with two hidden
layers of 64 units, ReLU, batch normalization, and dropout 0.3. CT,
PET-CT, and MRI: a compact 3D convolutional encoder that produces a
64-dimensional feature, concatenated with the clinical covariates,
followed by a small multilayer perceptron head with dropout 0.2.
Pathology: a multilayer perceptron on the 768-dimensional TITAN slide
feature, with layer normalization and hidden widths 256 and 128, GELU,
and dropout 0.3.

\textbf{Imaging preprocessing.} Volumes are cropped to a tumor-centered
cube of 128 millimeters guided by the segmentation and resampled to 64
by 64 by 64 voxels at 2 millimeter isotropic spacing. Magnetic resonance
sequences are z-scored within the brain region per sequence, computed
tomography intensities are windowed to a fixed Hounsfield range, and the
HECKTOR positron emission tomography and computed tomography inputs use
the provided normalization.

\textbf{Software.} PyTorch, with lifelines and scikit-survival for the
survival metrics and torchmtlr for multi-task logistic regression.

\subsection{B. Evaluation metrics}\label{b.-evaluation-metrics}

\textbf{Harrell concordance index.} The fraction of comparable pairs
(the earlier subject had an observed event) whose risks are correctly
ordered, with ties in the risk contributing one half.

\textbf{Truncated Uno concordance index.} The
inverse-probability-of-censoring-weighted concordance index, restricted
to comparable pairs whose case event time is at most the eightieth
percentile of the event times. The truncation is used because the
inverse-probability-of-censoring weights become unstable in the sparse
tail of the follow-up, which otherwise inflates the variance of the
estimate.

\textbf{Time-dependent area under the curve.} The cumulative/dynamic
area under the curve averaged over twenty evaluation times placed at the
tenth to ninetieth percentiles of the test event times.

\textbf{Integrated Brier score.} For the scalar-risk methods (sigmoid,
Cox, TripleSurv, and the hybrid) a Breslow baseline estimated on the
training fold turns the risk into a survival function, S(t\textbar x) =
exp(-H0(t) exp(risk)); the discrete-time methods (multi-task logistic
regression and DeepHit) use their native predicted survival curves. The
Brier score is integrated over a common grid of twenty time points
spanning the observed event times, and a lower value denotes better
calibration.

\subsection{C. Baseline
configurations}\label{c.-baseline-configurations}

All neural baselines share the backbone and optimization above; only the
loss differs.

\textbf{Cox / DeepSurv.} Negative Cox partial log-likelihood, with the
Breslow handling of tied event times.

\textbf{Multi-task logistic regression (MTLR).} Discrete-time negative
log-likelihood (torchmtlr) trained with weight regularization strength
one. For all loss-metric coupling results, the monitored MTLR validation
loss is the pure negative log-likelihood with the weight penalty
removed, so that the quantity compared against the concordance index
reflects data fit rather than the weight norm, consistently with the
other losses. The training objective is unchanged.

\textbf{DeepHit.} A discrete-time likelihood term plus an exponential
ranking term, with ranking weight 0.2 and smoothing 0.1.

\textbf{TripleSurv.} Reimplemented from the paper, since no public code
is available, as the weighted sum of a likelihood term, an exponential
time-adaptive pairwise ranking term, and a calibration term, with
weights 1.0, 1.0, and 0.5, ranking temperature 1.0, and time scale 1.0.
The survival term inside the likelihood is clamped to a small positive
floor before the logarithm, because one minus the cumulative softmax can
become a tiny negative number at the last time bin and otherwise
produces non-finite values for censored subjects.

\textbf{Random survival forest.} scikit-survival, one hundred trees,
minimum fifteen samples per leaf.

\textbf{Hybrid (diagnostic only).} The SCL plus one half of a
logistic-hazard negative log-likelihood anchor. It is reported to test
the theory, not as a competing method.

\hypertarget{tab:S1}{}\subsubsection{Table~S1. Per-dataset results on the ten tabular
cohorts.}\label{table-s1.-per-dataset-results-on-the-ten-tabular-cohorts.}

\textbf{Table S1a. METABRIC.}

{\def\LTcaptype{none} 
\begin{longtable}[]{@{}
  >{\raggedright\arraybackslash}p{(\linewidth - 10\tabcolsep) * \real{0.1667}}
  >{\raggedright\arraybackslash}p{(\linewidth - 10\tabcolsep) * \real{0.1667}}
  >{\raggedright\arraybackslash}p{(\linewidth - 10\tabcolsep) * \real{0.1667}}
  >{\raggedright\arraybackslash}p{(\linewidth - 10\tabcolsep) * \real{0.1667}}
  >{\raggedright\arraybackslash}p{(\linewidth - 10\tabcolsep) * \real{0.1667}}
  >{\raggedright\arraybackslash}p{(\linewidth - 10\tabcolsep) * \real{0.1667}}@{}}
\toprule\noalign{}
\begin{minipage}[b]{\linewidth}\raggedright
Method
\end{minipage} & \begin{minipage}[b]{\linewidth}\raggedright
Harrell C-index
\end{minipage} & \begin{minipage}[b]{\linewidth}\raggedright
Truncated Uno
\end{minipage} & \begin{minipage}[b]{\linewidth}\raggedright
Time-dependent AUC
\end{minipage} & \begin{minipage}[b]{\linewidth}\raggedright
Loss-metric rank correlation
\end{minipage} & \begin{minipage}[b]{\linewidth}\raggedright
Integrated Brier score
\end{minipage} \\
\midrule\noalign{}
\endhead
\bottomrule\noalign{}
\endlastfoot
Ours-Sigmoid & 0.725 \(\pm\) 0.000 & 0.716 \(\pm\) 0.003 & 0.765 \(\pm\)
0.005 & 0.982 \(\pm\) 0.006 & 0.163 \\
Ours-Hybrid & 0.695 \(\pm\) 0.006 & 0.687 \(\pm\) 0.002 & 0.740 \(\pm\)
0.006 & 0.243 \(\pm\) 0.093 & 0.472 \\
Cox & 0.733 \(\pm\) 0.002 & 0.724 \(\pm\) 0.005 & 0.770 \(\pm\) 0.006 &
0.096 \(\pm\) 0.158 & 0.161 \\
MTLR & 0.724 \(\pm\) 0.005 & 0.715 \(\pm\) 0.006 & 0.755 \(\pm\) 0.011 &
0.306 \(\pm\) 0.309 & 0.163 \\
DeepHit & 0.726 \(\pm\) 0.005 & 0.718 \(\pm\) 0.004 & 0.755 \(\pm\)
0.006 & -0.157 \(\pm\) 0.179 & 0.164 \\
TripleSurv & 0.649 \(\pm\) 0.006 & 0.640 \(\pm\) 0.005 & 0.676 \(\pm\)
0.003 & 0.523 \(\pm\) 0.074 & 0.180 \\
RSF & 0.791 \(\pm\) 0.002 & 0.779 \(\pm\) 0.002 & 0.817 \(\pm\) 0.002 &
n/a & n/a \\
\end{longtable}
}

\textbf{Table S1b. SUPPORT.}

{\def\LTcaptype{none} 
\begin{longtable}[]{@{}
  >{\raggedright\arraybackslash}p{(\linewidth - 10\tabcolsep) * \real{0.1667}}
  >{\raggedright\arraybackslash}p{(\linewidth - 10\tabcolsep) * \real{0.1667}}
  >{\raggedright\arraybackslash}p{(\linewidth - 10\tabcolsep) * \real{0.1667}}
  >{\raggedright\arraybackslash}p{(\linewidth - 10\tabcolsep) * \real{0.1667}}
  >{\raggedright\arraybackslash}p{(\linewidth - 10\tabcolsep) * \real{0.1667}}
  >{\raggedright\arraybackslash}p{(\linewidth - 10\tabcolsep) * \real{0.1667}}@{}}
\toprule\noalign{}
\begin{minipage}[b]{\linewidth}\raggedright
Method
\end{minipage} & \begin{minipage}[b]{\linewidth}\raggedright
Harrell C-index
\end{minipage} & \begin{minipage}[b]{\linewidth}\raggedright
Truncated Uno
\end{minipage} & \begin{minipage}[b]{\linewidth}\raggedright
Time-dependent AUC
\end{minipage} & \begin{minipage}[b]{\linewidth}\raggedright
Loss-metric rank correlation
\end{minipage} & \begin{minipage}[b]{\linewidth}\raggedright
Integrated Brier score
\end{minipage} \\
\midrule\noalign{}
\endhead
\bottomrule\noalign{}
\endlastfoot
Ours-Sigmoid & 0.886 \(\pm\) 0.001 & 0.907 \(\pm\) 0.001 & 0.966 \(\pm\)
0.000 & 0.995 & 0.133 \\
Ours-Hybrid & 0.892 \(\pm\) 0.001 & 0.914 \(\pm\) 0.000 & 0.969 \(\pm\)
0.000 & 0.265 \(\pm\) 0.153 & 0.092 \\
Cox & 0.889 \(\pm\) 0.001 & 0.910 \(\pm\) 0.001 & 0.967 \(\pm\) 0.000 &
0.220 \(\pm\) 0.079 & 0.126 \\
MTLR & 0.886 \(\pm\) 0.001 & 0.907 \(\pm\) 0.001 & 0.966 \(\pm\) 0.000 &
-0.823 \(\pm\) 0.021 & 0.092 \\
DeepHit & 0.888 \(\pm\) 0.001 & 0.910 \(\pm\) 0.001 & 0.967 & 0.225
\(\pm\) 0.070 & 0.103 \\
TripleSurv & 0.846 \(\pm\) 0.002 & 0.866 \(\pm\) 0.002 & 0.938 \(\pm\)
0.003 & n/a & 0.127 \\
RSF & 0.821 \(\pm\) 0.001 & 0.836 \(\pm\) 0.001 & 0.908 \(\pm\) 0.001 &
n/a & n/a \\
\end{longtable}
}

\textbf{Table S1c. GBSG.}

{\def\LTcaptype{none} 
\begin{longtable}[]{@{}
  >{\raggedright\arraybackslash}p{(\linewidth - 10\tabcolsep) * \real{0.1667}}
  >{\raggedright\arraybackslash}p{(\linewidth - 10\tabcolsep) * \real{0.1667}}
  >{\raggedright\arraybackslash}p{(\linewidth - 10\tabcolsep) * \real{0.1667}}
  >{\raggedright\arraybackslash}p{(\linewidth - 10\tabcolsep) * \real{0.1667}}
  >{\raggedright\arraybackslash}p{(\linewidth - 10\tabcolsep) * \real{0.1667}}
  >{\raggedright\arraybackslash}p{(\linewidth - 10\tabcolsep) * \real{0.1667}}@{}}
\toprule\noalign{}
\begin{minipage}[b]{\linewidth}\raggedright
Method
\end{minipage} & \begin{minipage}[b]{\linewidth}\raggedright
Harrell C-index
\end{minipage} & \begin{minipage}[b]{\linewidth}\raggedright
Truncated Uno
\end{minipage} & \begin{minipage}[b]{\linewidth}\raggedright
Time-dependent AUC
\end{minipage} & \begin{minipage}[b]{\linewidth}\raggedright
Loss-metric rank correlation
\end{minipage} & \begin{minipage}[b]{\linewidth}\raggedright
Integrated Brier score
\end{minipage} \\
\midrule\noalign{}
\endhead
\bottomrule\noalign{}
\endlastfoot
Ours-Sigmoid & 0.659 \(\pm\) 0.014 & 0.664 \(\pm\) 0.015 & 0.711 \(\pm\)
0.015 & 0.960 \(\pm\) 0.010 & 0.178 \\
Ours-Hybrid & 0.656 \(\pm\) 0.011 & 0.660 \(\pm\) 0.011 & 0.710 \(\pm\)
0.012 & 0.069 \(\pm\) 0.078 & 0.451 \\
Cox & 0.658 \(\pm\) 0.014 & 0.662 \(\pm\) 0.014 & 0.707 \(\pm\) 0.016 &
0.640 \(\pm\) 0.156 & 0.175 \\
MTLR & 0.659 \(\pm\) 0.017 & 0.663 \(\pm\) 0.019 & 0.710 \(\pm\) 0.025 &
0.303 \(\pm\) 0.258 & 0.174 \\
DeepHit & 0.625 \(\pm\) 0.019 & 0.628 \(\pm\) 0.021 & 0.672 \(\pm\)
0.023 & 0.078 \(\pm\) 0.043 & 0.247 \\
TripleSurv & 0.603 \(\pm\) 0.026 & 0.607 \(\pm\) 0.029 & 0.638 \(\pm\)
0.041 & -0.001 \(\pm\) 0.398 & 0.180 \\
RSF & 0.676 \(\pm\) 0.004 & 0.676 \(\pm\) 0.005 & 0.735 \(\pm\) 0.006 &
n/a & n/a \\
\end{longtable}
}

\textbf{Table S1d. ROTTERDAM.}

{\def\LTcaptype{none} 
\begin{longtable}[]{@{}
  >{\raggedright\arraybackslash}p{(\linewidth - 10\tabcolsep) * \real{0.1667}}
  >{\raggedright\arraybackslash}p{(\linewidth - 10\tabcolsep) * \real{0.1667}}
  >{\raggedright\arraybackslash}p{(\linewidth - 10\tabcolsep) * \real{0.1667}}
  >{\raggedright\arraybackslash}p{(\linewidth - 10\tabcolsep) * \real{0.1667}}
  >{\raggedright\arraybackslash}p{(\linewidth - 10\tabcolsep) * \real{0.1667}}
  >{\raggedright\arraybackslash}p{(\linewidth - 10\tabcolsep) * \real{0.1667}}@{}}
\toprule\noalign{}
\begin{minipage}[b]{\linewidth}\raggedright
Method
\end{minipage} & \begin{minipage}[b]{\linewidth}\raggedright
Harrell C-index
\end{minipage} & \begin{minipage}[b]{\linewidth}\raggedright
Truncated Uno
\end{minipage} & \begin{minipage}[b]{\linewidth}\raggedright
Time-dependent AUC
\end{minipage} & \begin{minipage}[b]{\linewidth}\raggedright
Loss-metric rank correlation
\end{minipage} & \begin{minipage}[b]{\linewidth}\raggedright
Integrated Brier score
\end{minipage} \\
\midrule\noalign{}
\endhead
\bottomrule\noalign{}
\endlastfoot
Ours-Sigmoid & 0.709 \(\pm\) 0.001 & 0.713 \(\pm\) 0.001 & 0.754 \(\pm\)
0.002 & 0.986 \(\pm\) 0.001 & 0.160 \\
Ours-Hybrid & 0.709 \(\pm\) 0.001 & 0.713 \(\pm\) 0.001 & 0.755 \(\pm\)
0.001 & 0.403 \(\pm\) 0.169 & 0.251 \\
Cox & 0.707 \(\pm\) 0.002 & 0.710 \(\pm\) 0.002 & 0.748 \(\pm\) 0.003 &
0.646 \(\pm\) 0.192 & 0.156 \\
MTLR & 0.708 \(\pm\) 0.001 & 0.712 \(\pm\) 0.001 & 0.756 \(\pm\) 0.001 &
0.629 \(\pm\) 0.157 & 0.153 \\
DeepHit & 0.659 \(\pm\) 0.011 & 0.663 \(\pm\) 0.010 & 0.699 \(\pm\)
0.012 & 0.079 \(\pm\) 0.108 & 0.228 \\
TripleSurv & 0.673 \(\pm\) 0.004 & 0.678 \(\pm\) 0.005 & 0.715 \(\pm\)
0.005 & n/a & 0.162 \\
RSF & 0.714 \(\pm\) 0.001 & 0.718 \(\pm\) 0.001 & 0.760 \(\pm\) 0.002 &
n/a & n/a \\
\end{longtable}
}

\textbf{Table S1e. FLCHAIN.}

{\def\LTcaptype{none} 
\begin{longtable}[]{@{}
  >{\raggedright\arraybackslash}p{(\linewidth - 10\tabcolsep) * \real{0.1667}}
  >{\raggedright\arraybackslash}p{(\linewidth - 10\tabcolsep) * \real{0.1667}}
  >{\raggedright\arraybackslash}p{(\linewidth - 10\tabcolsep) * \real{0.1667}}
  >{\raggedright\arraybackslash}p{(\linewidth - 10\tabcolsep) * \real{0.1667}}
  >{\raggedright\arraybackslash}p{(\linewidth - 10\tabcolsep) * \real{0.1667}}
  >{\raggedright\arraybackslash}p{(\linewidth - 10\tabcolsep) * \real{0.1667}}@{}}
\toprule\noalign{}
\begin{minipage}[b]{\linewidth}\raggedright
Method
\end{minipage} & \begin{minipage}[b]{\linewidth}\raggedright
Harrell C-index
\end{minipage} & \begin{minipage}[b]{\linewidth}\raggedright
Truncated Uno
\end{minipage} & \begin{minipage}[b]{\linewidth}\raggedright
Time-dependent AUC
\end{minipage} & \begin{minipage}[b]{\linewidth}\raggedright
Loss-metric rank correlation
\end{minipage} & \begin{minipage}[b]{\linewidth}\raggedright
Integrated Brier score
\end{minipage} \\
\midrule\noalign{}
\endhead
\bottomrule\noalign{}
\endlastfoot
Ours-Sigmoid & 0.934 & 0.929 \(\pm\) 0.000 & 0.948 \(\pm\) 0.000 & 0.988
\(\pm\) 0.004 & 0.068 \\
Ours-Hybrid & 0.933 \(\pm\) 0.001 & 0.927 \(\pm\) 0.001 & 0.945 \(\pm\)
0.000 & 0.108 \(\pm\) 0.114 & 0.059 \\
Cox & 0.931 \(\pm\) 0.002 & 0.925 \(\pm\) 0.002 & 0.943 \(\pm\) 0.002 &
0.677 \(\pm\) 0.038 & 0.061 \\
MTLR & 0.933 \(\pm\) 0.000 & 0.928 \(\pm\) 0.001 & 0.946 \(\pm\) 0.001 &
-0.285 \(\pm\) 0.079 & 0.053 \\
DeepHit & 0.933 \(\pm\) 0.001 & 0.928 \(\pm\) 0.001 & 0.946 \(\pm\)
0.001 & -0.003 \(\pm\) 0.122 & 0.098 \\
TripleSurv & 0.577 \(\pm\) 0.064 & 0.585 \(\pm\) 0.060 & 0.604 \(\pm\)
0.066 & n/a & 0.125 \\
RSF & 0.933 \(\pm\) 0.000 & 0.928 & 0.946 \(\pm\) 0.000 & n/a & n/a \\
\end{longtable}
}

\textbf{Table S1f. WHAS500.}

{\def\LTcaptype{none} 
\begin{longtable}[]{@{}
  >{\raggedright\arraybackslash}p{(\linewidth - 10\tabcolsep) * \real{0.1667}}
  >{\raggedright\arraybackslash}p{(\linewidth - 10\tabcolsep) * \real{0.1667}}
  >{\raggedright\arraybackslash}p{(\linewidth - 10\tabcolsep) * \real{0.1667}}
  >{\raggedright\arraybackslash}p{(\linewidth - 10\tabcolsep) * \real{0.1667}}
  >{\raggedright\arraybackslash}p{(\linewidth - 10\tabcolsep) * \real{0.1667}}
  >{\raggedright\arraybackslash}p{(\linewidth - 10\tabcolsep) * \real{0.1667}}@{}}
\toprule\noalign{}
\begin{minipage}[b]{\linewidth}\raggedright
Method
\end{minipage} & \begin{minipage}[b]{\linewidth}\raggedright
Harrell C-index
\end{minipage} & \begin{minipage}[b]{\linewidth}\raggedright
Truncated Uno
\end{minipage} & \begin{minipage}[b]{\linewidth}\raggedright
Time-dependent AUC
\end{minipage} & \begin{minipage}[b]{\linewidth}\raggedright
Loss-metric rank correlation
\end{minipage} & \begin{minipage}[b]{\linewidth}\raggedright
Integrated Brier score
\end{minipage} \\
\midrule\noalign{}
\endhead
\bottomrule\noalign{}
\endlastfoot
Ours-Sigmoid & 0.749 \(\pm\) 0.008 & 0.751 \(\pm\) 0.007 & 0.773 \(\pm\)
0.007 & 0.957 \(\pm\) 0.004 & 0.161 \\
Ours-Hybrid & 0.743 \(\pm\) 0.007 & 0.746 \(\pm\) 0.007 & 0.769 \(\pm\)
0.008 & -0.336 \(\pm\) 0.271 & 0.510 \\
Cox & 0.742 \(\pm\) 0.008 & 0.746 \(\pm\) 0.007 & 0.764 \(\pm\) 0.010 &
0.498 \(\pm\) 0.212 & 0.161 \\
MTLR & 0.748 \(\pm\) 0.000 & 0.753 \(\pm\) 0.000 & 0.782 \(\pm\) 0.003 &
0.100 \(\pm\) 0.477 & 0.171 \\
DeepHit & 0.718 \(\pm\) 0.018 & 0.722 \(\pm\) 0.018 & 0.740 \(\pm\)
0.022 & 0.372 \(\pm\) 0.060 & 0.254 \\
TripleSurv & 0.689 \(\pm\) 0.022 & 0.690 \(\pm\) 0.024 & 0.708 \(\pm\)
0.020 & 0.412 \(\pm\) 0.075 & 0.169 \\
RSF & 0.764 \(\pm\) 0.004 & 0.768 \(\pm\) 0.005 & 0.785 \(\pm\) 0.002 &
n/a & n/a \\
\end{longtable}
}

\textbf{Table S1g. SEER.}

{\def\LTcaptype{none} 
\begin{longtable}[]{@{}
  >{\raggedright\arraybackslash}p{(\linewidth - 10\tabcolsep) * \real{0.1667}}
  >{\raggedright\arraybackslash}p{(\linewidth - 10\tabcolsep) * \real{0.1667}}
  >{\raggedright\arraybackslash}p{(\linewidth - 10\tabcolsep) * \real{0.1667}}
  >{\raggedright\arraybackslash}p{(\linewidth - 10\tabcolsep) * \real{0.1667}}
  >{\raggedright\arraybackslash}p{(\linewidth - 10\tabcolsep) * \real{0.1667}}
  >{\raggedright\arraybackslash}p{(\linewidth - 10\tabcolsep) * \real{0.1667}}@{}}
\toprule\noalign{}
\begin{minipage}[b]{\linewidth}\raggedright
Method
\end{minipage} & \begin{minipage}[b]{\linewidth}\raggedright
Harrell C-index
\end{minipage} & \begin{minipage}[b]{\linewidth}\raggedright
Truncated Uno
\end{minipage} & \begin{minipage}[b]{\linewidth}\raggedright
Time-dependent AUC
\end{minipage} & \begin{minipage}[b]{\linewidth}\raggedright
Loss-metric rank correlation
\end{minipage} & \begin{minipage}[b]{\linewidth}\raggedright
Integrated Brier score
\end{minipage} \\
\midrule\noalign{}
\endhead
\bottomrule\noalign{}
\endlastfoot
Ours-Sigmoid & 0.721 \(\pm\) 0.006 & 0.723 \(\pm\) 0.006 & 0.741 \(\pm\)
0.006 & 0.996 \(\pm\) 0.001 & 0.072 \\
Ours-Hybrid & 0.720 \(\pm\) 0.005 & 0.721 \(\pm\) 0.004 & 0.742 \(\pm\)
0.004 & 0.817 \(\pm\) 0.074 & 0.353 \\
Cox & 0.719 \(\pm\) 0.001 & 0.722 \(\pm\) 0.001 & 0.740 \(\pm\) 0.001 &
0.726 \(\pm\) 0.049 & 0.071 \\
MTLR & 0.705 \(\pm\) 0.018 & 0.703 \(\pm\) 0.019 & 0.732 \(\pm\) 0.016 &
0.254 \(\pm\) 0.108 & 0.074 \\
DeepHit & 0.630 \(\pm\) 0.026 & 0.629 \(\pm\) 0.026 & 0.641 \(\pm\)
0.034 & -0.057 \(\pm\) 0.197 & 0.133 \\
TripleSurv & 0.533 \(\pm\) 0.011 & 0.535 \(\pm\) 0.011 & 0.535 \(\pm\)
0.010 & n/a & 0.080 \\
RSF & 0.724 \(\pm\) 0.001 & 0.725 \(\pm\) 0.001 & 0.745 \(\pm\) 0.001 &
n/a & n/a \\
\end{longtable}
}

\textbf{Table S1h. PBC.}

{\def\LTcaptype{none} 
\begin{longtable}[]{@{}
  >{\raggedright\arraybackslash}p{(\linewidth - 10\tabcolsep) * \real{0.1667}}
  >{\raggedright\arraybackslash}p{(\linewidth - 10\tabcolsep) * \real{0.1667}}
  >{\raggedright\arraybackslash}p{(\linewidth - 10\tabcolsep) * \real{0.1667}}
  >{\raggedright\arraybackslash}p{(\linewidth - 10\tabcolsep) * \real{0.1667}}
  >{\raggedright\arraybackslash}p{(\linewidth - 10\tabcolsep) * \real{0.1667}}
  >{\raggedright\arraybackslash}p{(\linewidth - 10\tabcolsep) * \real{0.1667}}@{}}
\toprule\noalign{}
\begin{minipage}[b]{\linewidth}\raggedright
Method
\end{minipage} & \begin{minipage}[b]{\linewidth}\raggedright
Harrell C-index
\end{minipage} & \begin{minipage}[b]{\linewidth}\raggedright
Truncated Uno
\end{minipage} & \begin{minipage}[b]{\linewidth}\raggedright
Time-dependent AUC
\end{minipage} & \begin{minipage}[b]{\linewidth}\raggedright
Loss-metric rank correlation
\end{minipage} & \begin{minipage}[b]{\linewidth}\raggedright
Integrated Brier score
\end{minipage} \\
\midrule\noalign{}
\endhead
\bottomrule\noalign{}
\endlastfoot
Ours-Sigmoid & 0.972 \(\pm\) 0.002 & 0.974 \(\pm\) 0.002 & 0.993 \(\pm\)
0.001 & 0.931 \(\pm\) 0.014 & 0.061 \\
Ours-Hybrid & 0.974 \(\pm\) 0.002 & 0.976 \(\pm\) 0.002 & 0.993 \(\pm\)
0.001 & 0.794 \(\pm\) 0.017 & 0.047 \\
Cox & 0.979 \(\pm\) 0.003 & 0.980 \(\pm\) 0.003 & 0.996 \(\pm\) 0.001 &
0.651 \(\pm\) 0.116 & 0.059 \\
MTLR & 0.963 \(\pm\) 0.004 & 0.964 \(\pm\) 0.003 & 0.987 \(\pm\) 0.002 &
0.932 \(\pm\) 0.023 & 0.039 \\
DeepHit & 0.950 \(\pm\) 0.002 & 0.952 \(\pm\) 0.004 & 0.978 \(\pm\)
0.002 & 0.335 \(\pm\) 0.130 & 0.065 \\
TripleSurv & 0.911 \(\pm\) 0.008 & 0.914 \(\pm\) 0.010 & 0.952 \(\pm\)
0.007 & 0.027 \(\pm\) 0.105 & 0.086 \\
RSF & 0.959 \(\pm\) 0.002 & 0.953 \(\pm\) 0.002 & 0.977 \(\pm\) 0.002 &
n/a & n/a \\
\end{longtable}
}

\textbf{Table S1i. lung.}

{\def\LTcaptype{none} 
\begin{longtable}[]{@{}
  >{\raggedright\arraybackslash}p{(\linewidth - 10\tabcolsep) * \real{0.1667}}
  >{\raggedright\arraybackslash}p{(\linewidth - 10\tabcolsep) * \real{0.1667}}
  >{\raggedright\arraybackslash}p{(\linewidth - 10\tabcolsep) * \real{0.1667}}
  >{\raggedright\arraybackslash}p{(\linewidth - 10\tabcolsep) * \real{0.1667}}
  >{\raggedright\arraybackslash}p{(\linewidth - 10\tabcolsep) * \real{0.1667}}
  >{\raggedright\arraybackslash}p{(\linewidth - 10\tabcolsep) * \real{0.1667}}@{}}
\toprule\noalign{}
\begin{minipage}[b]{\linewidth}\raggedright
Method
\end{minipage} & \begin{minipage}[b]{\linewidth}\raggedright
Harrell C-index
\end{minipage} & \begin{minipage}[b]{\linewidth}\raggedright
Truncated Uno
\end{minipage} & \begin{minipage}[b]{\linewidth}\raggedright
Time-dependent AUC
\end{minipage} & \begin{minipage}[b]{\linewidth}\raggedright
Loss-metric rank correlation
\end{minipage} & \begin{minipage}[b]{\linewidth}\raggedright
Integrated Brier score
\end{minipage} \\
\midrule\noalign{}
\endhead
\bottomrule\noalign{}
\endlastfoot
Ours-Sigmoid & 0.611 \(\pm\) 0.015 & 0.607 \(\pm\) 0.015 & 0.623 \(\pm\)
0.020 & 0.918 \(\pm\) 0.023 & 0.220 \\
Ours-Hybrid & 0.606 \(\pm\) 0.010 & 0.603 \(\pm\) 0.013 & 0.616 \(\pm\)
0.016 & -0.153 \(\pm\) 0.180 & 0.344 \\
Cox & 0.589 \(\pm\) 0.014 & 0.592 \(\pm\) 0.015 & 0.598 \(\pm\) 0.023 &
0.358 \(\pm\) 0.087 & 0.228 \\
MTLR & 0.615 \(\pm\) 0.013 & 0.610 \(\pm\) 0.012 & 0.631 \(\pm\) 0.024 &
0.166 \(\pm\) 0.356 & 0.204 \\
DeepHit & 0.602 \(\pm\) 0.008 & 0.598 \(\pm\) 0.009 & 0.614 \(\pm\)
0.012 & 0.225 \(\pm\) 0.112 & 0.210 \\
TripleSurv & 0.594 \(\pm\) 0.024 & 0.589 \(\pm\) 0.026 & 0.603 \(\pm\)
0.035 & 0.056 \(\pm\) 0.274 & 0.218 \\
RSF & 0.628 \(\pm\) 0.005 & 0.622 \(\pm\) 0.006 & 0.649 \(\pm\) 0.007 &
n/a & n/a \\
\end{longtable}
}

\textbf{Table S1j. TCGA-GBM.}

{\def\LTcaptype{none} 
\begin{longtable}[]{@{}
  >{\raggedright\arraybackslash}p{(\linewidth - 10\tabcolsep) * \real{0.1667}}
  >{\raggedright\arraybackslash}p{(\linewidth - 10\tabcolsep) * \real{0.1667}}
  >{\raggedright\arraybackslash}p{(\linewidth - 10\tabcolsep) * \real{0.1667}}
  >{\raggedright\arraybackslash}p{(\linewidth - 10\tabcolsep) * \real{0.1667}}
  >{\raggedright\arraybackslash}p{(\linewidth - 10\tabcolsep) * \real{0.1667}}
  >{\raggedright\arraybackslash}p{(\linewidth - 10\tabcolsep) * \real{0.1667}}@{}}
\toprule\noalign{}
\begin{minipage}[b]{\linewidth}\raggedright
Method
\end{minipage} & \begin{minipage}[b]{\linewidth}\raggedright
Harrell C-index
\end{minipage} & \begin{minipage}[b]{\linewidth}\raggedright
Truncated Uno
\end{minipage} & \begin{minipage}[b]{\linewidth}\raggedright
Time-dependent AUC
\end{minipage} & \begin{minipage}[b]{\linewidth}\raggedright
Loss-metric rank correlation
\end{minipage} & \begin{minipage}[b]{\linewidth}\raggedright
Integrated Brier score
\end{minipage} \\
\midrule\noalign{}
\endhead
\bottomrule\noalign{}
\endlastfoot
Ours-Sigmoid & 0.879 \(\pm\) 0.004 & 0.879 \(\pm\) 0.004 & 0.936 \(\pm\)
0.002 & 0.890 \(\pm\) 0.026 & 0.095 \\
Ours-Hybrid & 0.877 \(\pm\) 0.001 & 0.876 \(\pm\) 0.001 & 0.936 \(\pm\)
0.001 & -0.042 \(\pm\) 0.240 & 0.166 \\
Cox & 0.862 \(\pm\) 0.001 & 0.860 \(\pm\) 0.001 & 0.924 \(\pm\) 0.002 &
0.445 \(\pm\) 0.049 & 0.102 \\
MTLR & 0.866 \(\pm\) 0.006 & 0.866 \(\pm\) 0.005 & 0.928 \(\pm\) 0.006 &
0.500 \(\pm\) 0.239 & 0.078 \\
DeepHit & 0.869 \(\pm\) 0.002 & 0.868 \(\pm\) 0.003 & 0.929 \(\pm\)
0.002 & 0.197 \(\pm\) 0.165 & 0.095 \\
TripleSurv & 0.855 \(\pm\) 0.004 & 0.854 \(\pm\) 0.004 & 0.921 \(\pm\)
0.006 & 0.262 \(\pm\) 0.163 & 0.113 \\
RSF & 0.853 \(\pm\) 0.004 & 0.853 \(\pm\) 0.004 & 0.920 \(\pm\) 0.002 &
n/a & n/a \\
\end{longtable}
}

\hypertarget{tab:S2}{}\subsubsection{Table~S2. Per-cohort results on the four TCGA pathology
cohorts.}\label{table-s2.-per-cohort-results-on-the-four-tcga-pathology-cohorts.}

\textbf{Table S2a. TCGA-BRCA.}

{\def\LTcaptype{none} 
\begin{longtable}[]{@{}
  >{\raggedright\arraybackslash}p{(\linewidth - 12\tabcolsep) * \real{0.1429}}
  >{\raggedright\arraybackslash}p{(\linewidth - 12\tabcolsep) * \real{0.1429}}
  >{\raggedright\arraybackslash}p{(\linewidth - 12\tabcolsep) * \real{0.1429}}
  >{\raggedright\arraybackslash}p{(\linewidth - 12\tabcolsep) * \real{0.1429}}
  >{\raggedright\arraybackslash}p{(\linewidth - 12\tabcolsep) * \real{0.1429}}
  >{\raggedright\arraybackslash}p{(\linewidth - 12\tabcolsep) * \real{0.1429}}
  >{\raggedright\arraybackslash}p{(\linewidth - 12\tabcolsep) * \real{0.1429}}@{}}
\toprule\noalign{}
\begin{minipage}[b]{\linewidth}\raggedright
Method
\end{minipage} & \begin{minipage}[b]{\linewidth}\raggedright
Harrell C-index
\end{minipage} & \begin{minipage}[b]{\linewidth}\raggedright
Truncated Uno
\end{minipage} & \begin{minipage}[b]{\linewidth}\raggedright
Time-dependent AUC
\end{minipage} & \begin{minipage}[b]{\linewidth}\raggedright
Integrated Brier score
\end{minipage} & \begin{minipage}[b]{\linewidth}\raggedright
Loss-metric rank correlation
\end{minipage} & \begin{minipage}[b]{\linewidth}\raggedright
Selection regret
\end{minipage} \\
\midrule\noalign{}
\endhead
\bottomrule\noalign{}
\endlastfoot
Ours-Sigmoid & 0.629 \(\pm\) 0.029 & 0.609 \(\pm\) 0.019 & 0.618 \(\pm\)
0.026 & 0.144 & 0.989 \(\pm\) 0.004 & -0.000 \(\pm\) 0.000 \\
Ours-Hybrid & 0.643 \(\pm\) 0.004 & 0.602 \(\pm\) 0.022 & 0.623 \(\pm\)
0.007 & 0.149 & 0.371 \(\pm\) 0.027 & -0.006 \(\pm\) 0.025 \\
Cox & 0.626 \(\pm\) 0.028 & 0.599 \(\pm\) 0.036 & 0.611 \(\pm\) 0.045 &
0.144 & 0.138 \(\pm\) 0.092 & -0.011 \(\pm\) 0.016 \\
MTLR & 0.559 \(\pm\) 0.020 & 0.571 \(\pm\) 0.008 & 0.566 \(\pm\) 0.016 &
0.146 & 0.233 \(\pm\) 0.216 & 0.023 \(\pm\) 0.014 \\
DeepHit & 0.620 \(\pm\) 0.017 & 0.617 \(\pm\) 0.018 & 0.606 \(\pm\)
0.024 & 0.142 & 0.238 \(\pm\) 0.089 & -0.011 \(\pm\) 0.038 \\
TripleSurv & 0.554 \(\pm\) 0.017 & 0.544 \(\pm\) 0.009 & 0.567 \(\pm\)
0.006 & 0.143 & -0.101 \(\pm\) 0.081 & 0.064 \(\pm\) 0.020 \\
\end{longtable}
}

\textbf{Table S2b. TCGA-GBMLGG.}

{\def\LTcaptype{none} 
\begin{longtable}[]{@{}
  >{\raggedright\arraybackslash}p{(\linewidth - 12\tabcolsep) * \real{0.1429}}
  >{\raggedright\arraybackslash}p{(\linewidth - 12\tabcolsep) * \real{0.1429}}
  >{\raggedright\arraybackslash}p{(\linewidth - 12\tabcolsep) * \real{0.1429}}
  >{\raggedright\arraybackslash}p{(\linewidth - 12\tabcolsep) * \real{0.1429}}
  >{\raggedright\arraybackslash}p{(\linewidth - 12\tabcolsep) * \real{0.1429}}
  >{\raggedright\arraybackslash}p{(\linewidth - 12\tabcolsep) * \real{0.1429}}
  >{\raggedright\arraybackslash}p{(\linewidth - 12\tabcolsep) * \real{0.1429}}@{}}
\toprule\noalign{}
\begin{minipage}[b]{\linewidth}\raggedright
Method
\end{minipage} & \begin{minipage}[b]{\linewidth}\raggedright
Harrell C-index
\end{minipage} & \begin{minipage}[b]{\linewidth}\raggedright
Truncated Uno
\end{minipage} & \begin{minipage}[b]{\linewidth}\raggedright
Time-dependent AUC
\end{minipage} & \begin{minipage}[b]{\linewidth}\raggedright
Integrated Brier score
\end{minipage} & \begin{minipage}[b]{\linewidth}\raggedright
Loss-metric rank correlation
\end{minipage} & \begin{minipage}[b]{\linewidth}\raggedright
Selection regret
\end{minipage} \\
\midrule\noalign{}
\endhead
\bottomrule\noalign{}
\endlastfoot
Ours-Sigmoid & 0.815 \(\pm\) 0.004 & 0.804 \(\pm\) 0.003 & 0.842 \(\pm\)
0.004 & 0.132 & 0.984 \(\pm\) 0.006 & 0.000 \\
Ours-Hybrid & 0.802 \(\pm\) 0.007 & 0.795 \(\pm\) 0.008 & 0.828 \(\pm\)
0.007 & 0.204 & 0.171 \(\pm\) 0.067 & -0.016 \(\pm\) 0.004 \\
Cox & 0.813 \(\pm\) 0.009 & 0.803 \(\pm\) 0.007 & 0.840 \(\pm\) 0.008 &
0.130 & 0.012 \(\pm\) 0.050 & -0.004 \(\pm\) 0.010 \\
MTLR & 0.809 \(\pm\) 0.012 & 0.798 \(\pm\) 0.009 & 0.835 \(\pm\) 0.010 &
0.132 & 0.161 \(\pm\) 0.072 & 0.026 \(\pm\) 0.006 \\
DeepHit & 0.799 \(\pm\) 0.007 & 0.789 \(\pm\) 0.007 & 0.821 \(\pm\)
0.011 & 0.156 & 0.257 \(\pm\) 0.035 & -0.019 \(\pm\) 0.010 \\
TripleSurv & 0.764 \(\pm\) 0.015 & 0.747 \(\pm\) 0.019 & 0.782 \(\pm\)
0.020 & 0.132 & 0.197 \(\pm\) 0.138 & 0.030 \(\pm\) 0.077 \\
\end{longtable}
}

\textbf{Table S2c. TCGA-LUAD.}

{\def\LTcaptype{none} 
\begin{longtable}[]{@{}
  >{\raggedright\arraybackslash}p{(\linewidth - 12\tabcolsep) * \real{0.1429}}
  >{\raggedright\arraybackslash}p{(\linewidth - 12\tabcolsep) * \real{0.1429}}
  >{\raggedright\arraybackslash}p{(\linewidth - 12\tabcolsep) * \real{0.1429}}
  >{\raggedright\arraybackslash}p{(\linewidth - 12\tabcolsep) * \real{0.1429}}
  >{\raggedright\arraybackslash}p{(\linewidth - 12\tabcolsep) * \real{0.1429}}
  >{\raggedright\arraybackslash}p{(\linewidth - 12\tabcolsep) * \real{0.1429}}
  >{\raggedright\arraybackslash}p{(\linewidth - 12\tabcolsep) * \real{0.1429}}@{}}
\toprule\noalign{}
\begin{minipage}[b]{\linewidth}\raggedright
Method
\end{minipage} & \begin{minipage}[b]{\linewidth}\raggedright
Harrell C-index
\end{minipage} & \begin{minipage}[b]{\linewidth}\raggedright
Truncated Uno
\end{minipage} & \begin{minipage}[b]{\linewidth}\raggedright
Time-dependent AUC
\end{minipage} & \begin{minipage}[b]{\linewidth}\raggedright
Integrated Brier score
\end{minipage} & \begin{minipage}[b]{\linewidth}\raggedright
Loss-metric rank correlation
\end{minipage} & \begin{minipage}[b]{\linewidth}\raggedright
Selection regret
\end{minipage} \\
\midrule\noalign{}
\endhead
\bottomrule\noalign{}
\endlastfoot
Ours-Sigmoid & 0.566 \(\pm\) 0.014 & 0.572 \(\pm\) 0.009 & 0.582 \(\pm\)
0.016 & 0.199 & 0.977 \(\pm\) 0.006 & -0.001 \(\pm\) 0.001 \\
Ours-Hybrid & 0.566 \(\pm\) 0.019 & 0.575 \(\pm\) 0.013 & 0.579 \(\pm\)
0.014 & 0.243 & -0.087 \(\pm\) 0.104 & -0.012 \(\pm\) 0.015 \\
Cox & 0.567 \(\pm\) 0.028 & 0.574 \(\pm\) 0.020 & 0.577 \(\pm\) 0.028 &
0.195 & 0.104 \(\pm\) 0.088 & -0.003 \(\pm\) 0.036 \\
MTLR & 0.549 \(\pm\) 0.027 & 0.554 \(\pm\) 0.026 & 0.553 \(\pm\) 0.028 &
0.202 & 0.127 \(\pm\) 0.093 & 0.014 \(\pm\) 0.022 \\
DeepHit & 0.553 \(\pm\) 0.005 & 0.559 \(\pm\) 0.015 & 0.555 \(\pm\)
0.009 & 0.203 & 0.121 \(\pm\) 0.035 & -0.021 \(\pm\) 0.014 \\
TripleSurv & 0.574 \(\pm\) 0.028 & 0.573 \(\pm\) 0.010 & 0.583 \(\pm\)
0.024 & 0.194 & -0.033 \(\pm\) 0.241 & 0.029 \(\pm\) 0.041 \\
\end{longtable}
}

\textbf{Table S2d. TCGA-UCEC.}

{\def\LTcaptype{none} 
\begin{longtable}[]{@{}
  >{\raggedright\arraybackslash}p{(\linewidth - 12\tabcolsep) * \real{0.1429}}
  >{\raggedright\arraybackslash}p{(\linewidth - 12\tabcolsep) * \real{0.1429}}
  >{\raggedright\arraybackslash}p{(\linewidth - 12\tabcolsep) * \real{0.1429}}
  >{\raggedright\arraybackslash}p{(\linewidth - 12\tabcolsep) * \real{0.1429}}
  >{\raggedright\arraybackslash}p{(\linewidth - 12\tabcolsep) * \real{0.1429}}
  >{\raggedright\arraybackslash}p{(\linewidth - 12\tabcolsep) * \real{0.1429}}
  >{\raggedright\arraybackslash}p{(\linewidth - 12\tabcolsep) * \real{0.1429}}@{}}
\toprule\noalign{}
\begin{minipage}[b]{\linewidth}\raggedright
Method
\end{minipage} & \begin{minipage}[b]{\linewidth}\raggedright
Harrell C-index
\end{minipage} & \begin{minipage}[b]{\linewidth}\raggedright
Truncated Uno
\end{minipage} & \begin{minipage}[b]{\linewidth}\raggedright
Time-dependent AUC
\end{minipage} & \begin{minipage}[b]{\linewidth}\raggedright
Integrated Brier score
\end{minipage} & \begin{minipage}[b]{\linewidth}\raggedright
Loss-metric rank correlation
\end{minipage} & \begin{minipage}[b]{\linewidth}\raggedright
Selection regret
\end{minipage} \\
\midrule\noalign{}
\endhead
\bottomrule\noalign{}
\endlastfoot
Ours-Sigmoid & 0.675 \(\pm\) 0.009 & 0.666 \(\pm\) 0.017 & 0.691 \(\pm\)
0.013 & 0.100 & 0.983 \(\pm\) 0.002 & -0.007 \(\pm\) 0.007 \\
Ours-Hybrid & 0.691 \(\pm\) 0.018 & 0.678 \(\pm\) 0.016 & 0.702 \(\pm\)
0.013 & 0.174 & 0.300 \(\pm\) 0.037 & -0.023 \(\pm\) 0.005 \\
Cox & 0.709 \(\pm\) 0.024 & 0.699 \(\pm\) 0.023 & 0.722 \(\pm\) 0.020 &
0.092 & 0.061 \(\pm\) 0.076 & -0.009 \(\pm\) 0.021 \\
MTLR & 0.684 \(\pm\) 0.026 & 0.675 \(\pm\) 0.033 & 0.693 \(\pm\) 0.027 &
0.104 & -0.038 \(\pm\) 0.185 & 0.045 \(\pm\) 0.019 \\
DeepHit & 0.597 \(\pm\) 0.041 & 0.579 \(\pm\) 0.038 & 0.592 \(\pm\)
0.050 & 0.151 & 0.415 \(\pm\) 0.232 & -0.055 \(\pm\) 0.030 \\
TripleSurv & 0.606 \(\pm\) 0.040 & 0.616 \(\pm\) 0.038 & 0.619 \(\pm\)
0.047 & 0.101 & 0.060 \(\pm\) 0.172 & -0.007 \(\pm\) 0.016 \\
\end{longtable}
}

\hypertarget{tab:S3}{}\subsubsection{Table~S3. Convex linear survival RankSVM versus the deep
SCL, per tabular
dataset.}\label{table-s3.-convex-linear-survival-ranksvm-versus-the-deep-scl-per-tabular-dataset.}

The convex linear counterpart of \hyperlink{sec:3.4}{Section 3.4} is solved to global
optimality with a quasi-Newton method under the same pooled five-fold
protocol. Harrell concordance index, mean plus or minus standard
deviation across three seeds.

{\def\LTcaptype{none} 
\begin{longtable}[]{@{}lll@{}}
\toprule\noalign{}
Dataset & Convex linear RankSVM & Deep SCL \\
\midrule\noalign{}
\endhead
\bottomrule\noalign{}
\endlastfoot
METABRIC & 0.734 \(\pm\) 0.003 & 0.725 \(\pm\) 0.000 \\
SUPPORT & 0.868 \(\pm\) 0.001 & 0.886 \(\pm\) 0.001 \\
GBSG & 0.678 \(\pm\) 0.004 & 0.659 \(\pm\) 0.014 \\
ROTTERDAM & 0.696 \(\pm\) 0.001 & 0.709 \(\pm\) 0.001 \\
FLCHAIN & 0.934 \(\pm\) 0.000 & 0.934 \\
WHAS500 & 0.766 \(\pm\) 0.002 & 0.749 \(\pm\) 0.008 \\
SEER & 0.727 \(\pm\) 0.005 & 0.721 \(\pm\) 0.006 \\
PBC & 0.995 \(\pm\) 0.000 & 0.972 \(\pm\) 0.002 \\
lung & 0.635 \(\pm\) 0.003 & 0.611 \(\pm\) 0.015 \\
TCGA-GBM & 0.885 \(\pm\) 0.003 & 0.879 \(\pm\) 0.004 \\
\textbf{Mean} & \textbf{0.792} & \textbf{0.785} \\
\end{longtable}
}

\hypertarget{tab:S4}{}\subsubsection{Table~S4. Pooled test C-index under checkpoint selection
by minimum validation
loss.}\label{table-s4.-pooled-test-c-index-under-checkpoint-selection-by-minimum-validation-loss.}

The same pooled five-fold protocol with three seeds as \hyperlink{tab:2}{Table 2}, but the
per-fold checkpoint is chosen by the minimum validation loss rather than
by the best validation concordance index. For the SCL the two rules give
the same test concordance index; for the likelihood and margin losses
selecting on the loss can lose concordance index (the selection regret
of \hyperlink{fig:5}{Figure 5}). The by-validation-concordance-index column reproduces the
main results up to cross-validation noise. Higher is better.

{\def\LTcaptype{none} 
\begin{longtable}[]{@{}
  >{\raggedright\arraybackslash}p{(\linewidth - 8\tabcolsep) * \real{0.2000}}
  >{\raggedright\arraybackslash}p{(\linewidth - 8\tabcolsep) * \real{0.2000}}
  >{\raggedright\arraybackslash}p{(\linewidth - 8\tabcolsep) * \real{0.2000}}
  >{\raggedright\arraybackslash}p{(\linewidth - 8\tabcolsep) * \real{0.2000}}
  >{\raggedright\arraybackslash}p{(\linewidth - 8\tabcolsep) * \real{0.2000}}@{}}
\toprule\noalign{}
\begin{minipage}[b]{\linewidth}\raggedright
Cohort
\end{minipage} & \begin{minipage}[b]{\linewidth}\raggedright
Method
\end{minipage} & \begin{minipage}[b]{\linewidth}\raggedright
C-index, select by val C-index
\end{minipage} & \begin{minipage}[b]{\linewidth}\raggedright
C-index, select by val loss
\end{minipage} & \begin{minipage}[b]{\linewidth}\raggedright
Regret
\end{minipage} \\
\midrule\noalign{}
\endhead
\bottomrule\noalign{}
\endlastfoot
HECKTOR (PET/CT) & \textbf{Ours-Sigmoid} & 0.644 & 0.655 & -0.011 \\
& Cox & 0.612 & 0.625 & -0.012 \\
& MTLR & 0.634 & 0.669 & -0.035 \\
& DeepHit & 0.596 & 0.616 & -0.020 \\
& TripleSurv & 0.566 & 0.552 & +0.014 \\
& Squared-hinge & 0.629 & 0.615 & +0.014 \\
CRLM (liver CT) & \textbf{Ours-Sigmoid} & 0.584 & 0.580 & +0.004 \\
& Cox & 0.600 & 0.595 & +0.005 \\
& MTLR & 0.575 & 0.594 & -0.019 \\
& DeepHit & 0.572 & 0.583 & -0.012 \\
& TripleSurv & 0.560 & 0.551 & +0.010 \\
& Squared-hinge & 0.586 & 0.604 & -0.018 \\
UCSF-PDGM (MRI) & \textbf{Ours-Sigmoid} & 0.711 & 0.708 & +0.003 \\
& Cox & 0.686 & 0.701 & -0.014 \\
& MTLR & 0.691 & 0.716 & -0.025 \\
& DeepHit & 0.675 & 0.713 & -0.038 \\
& TripleSurv & 0.632 & 0.596 & +0.037 \\
& Squared-hinge & 0.692 & 0.693 & -0.002 \\
UPENN-GBM (MRI) & \textbf{Ours-Sigmoid} & 0.631 & 0.637 & -0.006 \\
& Cox & 0.618 & 0.605 & +0.013 \\
& MTLR & 0.623 & 0.639 & -0.015 \\
& DeepHit & 0.638 & 0.637 & +0.001 \\
& TripleSurv & 0.628 & 0.575 & +0.053 \\
& Squared-hinge & 0.618 & 0.620 & -0.003 \\
Pathology (4 cohorts, mean) & \textbf{Ours-Sigmoid} & 0.681 & 0.680 &
+0.001 \\
& Cox & 0.677 & 0.674 & +0.003 \\
& MTLR & 0.660 & 0.664 & -0.003 \\
& DeepHit & 0.647 & 0.674 & -0.026 \\
& TripleSurv & 0.592 & 0.580 & +0.012 \\
& Squared-hinge & 0.670 & 0.675 & -0.005 \\
Tabular (10 datasets, mean) & \textbf{Ours-Sigmoid} & 0.785 & 0.785 &
-0.000 \\
& Cox & 0.781 & 0.780 & +0.001 \\
& MTLR & 0.781 & 0.781 & -0.000 \\
& DeepHit & 0.760 & 0.761 & -0.001 \\
& TripleSurv & 0.696 & 0.672 & +0.023 \\
& Squared-hinge & 0.780 & 0.777 & +0.003 \\
\end{longtable}
}

\hypertarget{fig:S1}{}\subsubsection{Figure~S1. Clinical risk
stratification.}\label{figure-s1.-clinical-risk-stratification.}

\begin{figure}
\hypertarget{fig:S1}{}\centering
\pandocbounded{\includegraphics[keepaspectratio,alt={\hyperlink{fig:S1}{Figure S1}. Kaplan-Meier survival curves for patients split at the median out-of-fold predicted risk of the SCL on the four imaging cohorts. The high-risk and low-risk groups separate significantly on every cohort by the log-rank test, with hazard ratios of 2.86 on HECKTOR for recurrence-free survival (95 per cent confidence interval 2.00 to 4.09), 2.82 on UCSF-PDGM (2.17 to 3.66), 1.69 on UPENN-GBM (1.43 to 1.99), and 1.87 on Colorectal-Liver-Metastases (1.27 to 2.75) for overall survival.}]{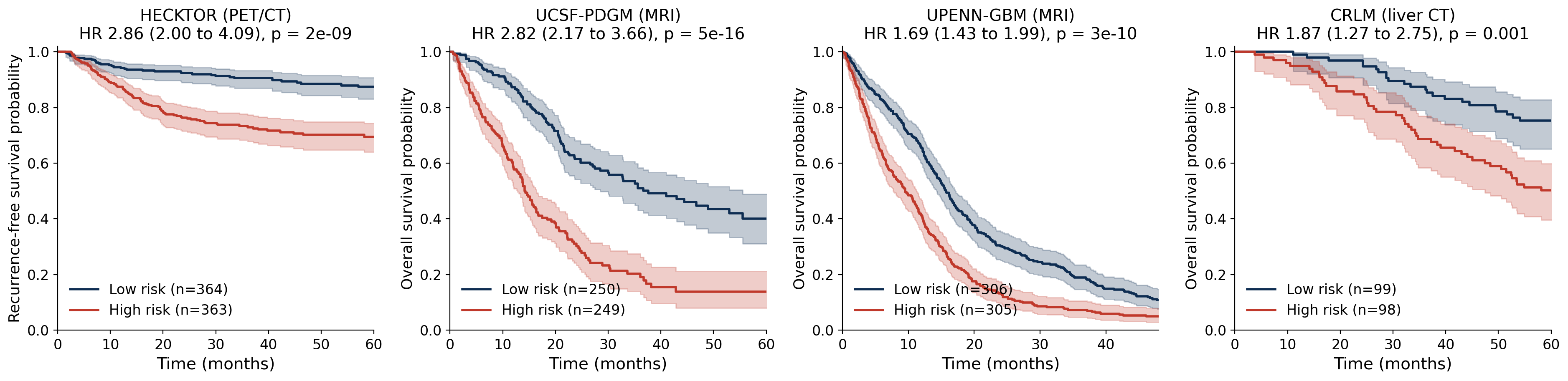}}
\caption{Figure~S1. Kaplan-Meier survival curves for patients split at
the median out-of-fold predicted risk of the SCL on the four imaging
cohorts. The high-risk and low-risk groups separate significantly on
every cohort by the log-rank test, with hazard ratios of 2.86 on HECKTOR
for recurrence-free survival (95 per cent confidence interval 2.00 to
4.09), 2.82 on UCSF-PDGM (2.17 to 3.66), 1.69 on UPENN-GBM (1.43 to
1.99), and 1.87 on Colorectal-Liver-Metastases (1.27 to 2.75) for
overall survival.}
\end{figure}

\hypertarget{fig:S2}{}\subsubsection{Figure~S2. Robustness to training
censoring.}\label{figure-s2.-robustness-to-training-censoring.}

\begin{figure}
\hypertarget{fig:S2}{}\centering
\pandocbounded{\includegraphics[keepaspectratio,alt={\hyperlink{fig:S2}{Figure S2}. Test concordance index as increasing synthetic censoring is added to the training data of two tabular cohorts (a fraction of the training subjects is randomly re-labeled as censored). The SCL degrades the least as censoring increases; DeepHit degrades the most on the lower-event-rate cohort (METABRIC). This is consistent with the loss being defined only over comparable pairs, which uses the censoring structure directly rather than modeling the full event distribution.}]{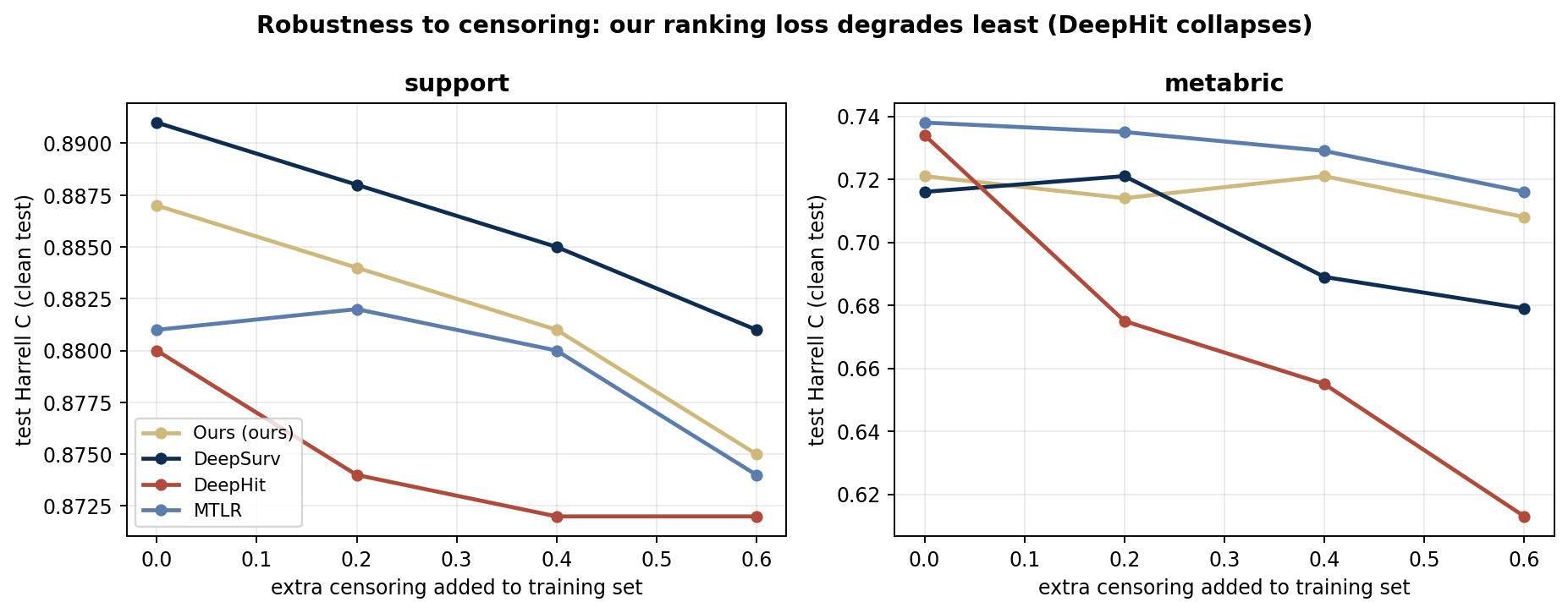}}
\caption{Figure~S2. Test concordance index as increasing synthetic
censoring is added to the training data of two tabular cohorts (a
fraction of the training subjects is randomly re-labeled as censored).
The SCL degrades the least as censoring increases; DeepHit degrades the
most on the lower-event-rate cohort (METABRIC). This is consistent with
the loss being defined only over comparable pairs, which uses the
censoring structure directly rather than modeling the full event
distribution.}
\end{figure}

\subsection{D. Loss-metric coupling on the remaining
cohorts}\label{d.-loss-metric-coupling-on-the-remaining-cohorts}

These figures repeat the training and validation trajectories of Figure
4 of the main text on the other cohorts. The dashed line is the data-fit
training loss, the solid line is the data-fit validation loss, and the
gold line is the validation concordance index; the gold star and navy
circle mark the checkpoint that would be selected by the best validation
concordance index and by the minimum validation loss. The sigmoid
concordance loss tracks the concordance index on every cohort, while the
likelihood and margin losses do not.

\begin{figure}
\hypertarget{fig:S3}{}\centering
\pandocbounded{\includegraphics[keepaspectratio,alt={\hyperlink{fig:S3}{Figure S3}. Loss-metric coupling on HECKTOR head-and-neck CT and PET (727 patients). Curves and markers are as in \hyperlink{fig:4}{Figure 4} of the main text.}]{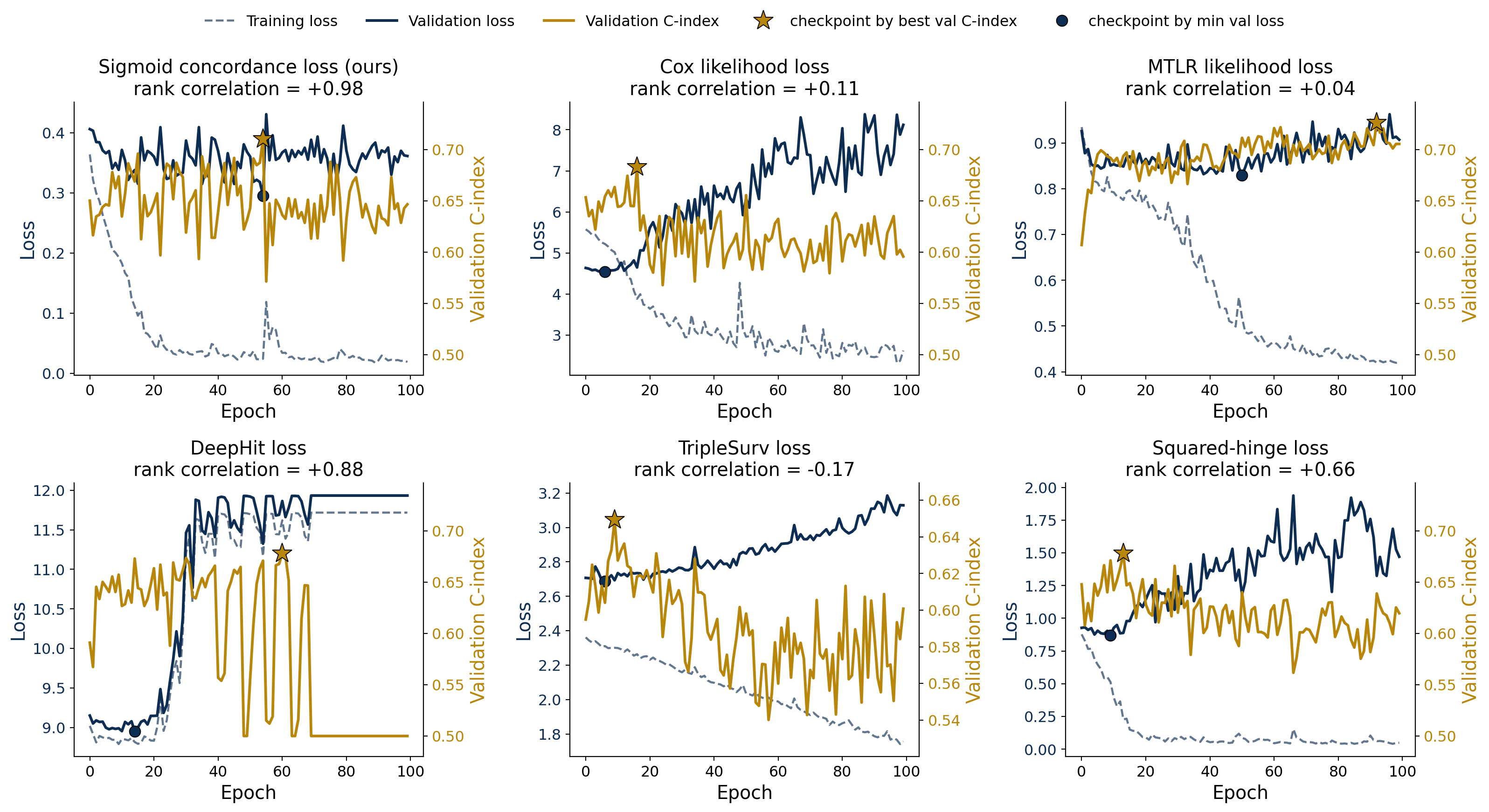}}
\caption{Figure~S3. Loss-metric coupling on HECKTOR head-and-neck CT and
PET (727 patients). Curves and markers are as in Figure 4 of the main
text.}
\end{figure}

\begin{figure}
\hypertarget{fig:S4}{}\centering
\pandocbounded{\includegraphics[keepaspectratio,alt={\hyperlink{fig:S4}{Figure S4}. Loss-metric coupling on UPENN-GBM MRI (611 patients). Curves and markers are as in \hyperlink{fig:4}{Figure 4} of the main text.}]{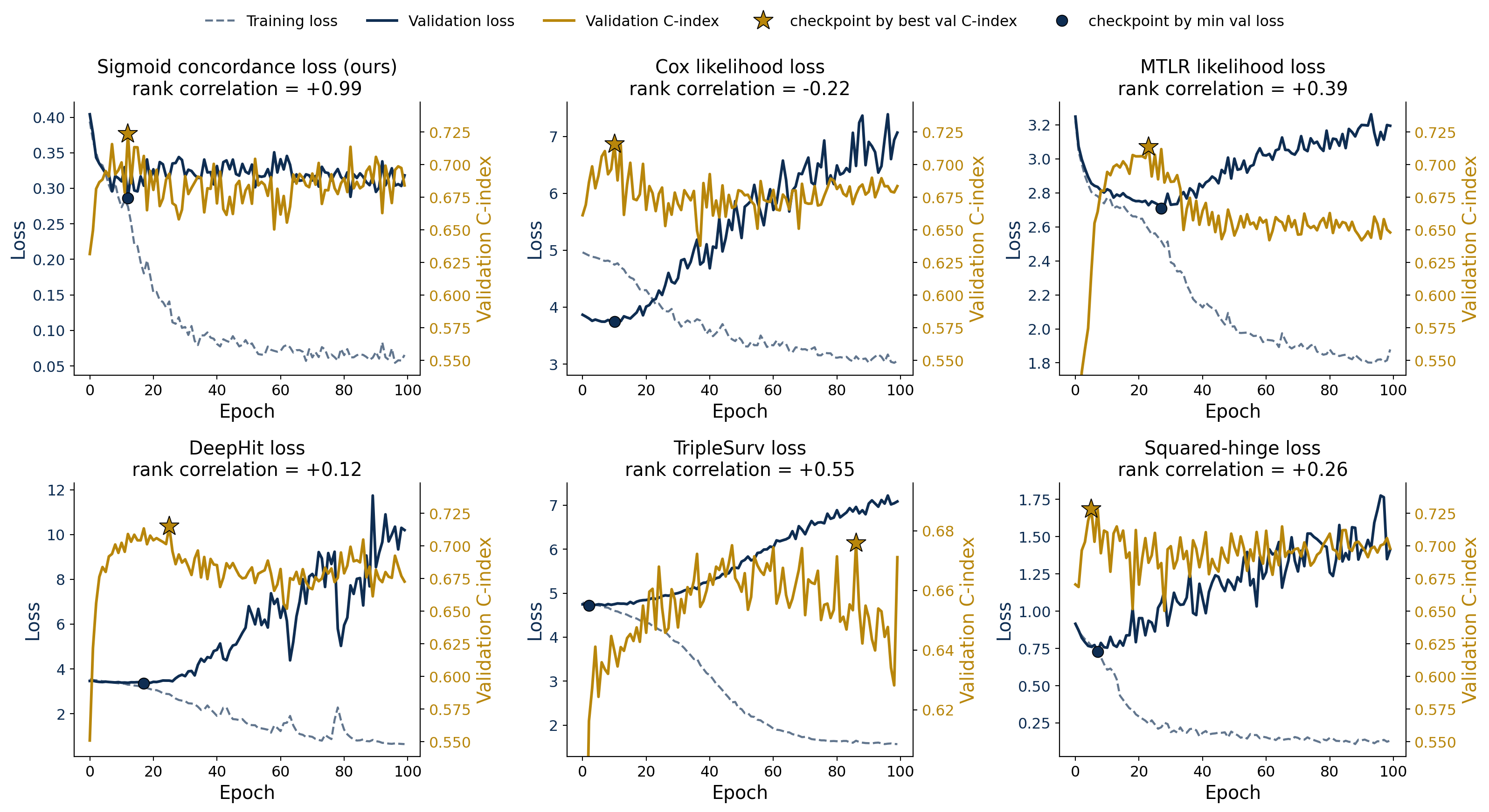}}
\caption{Figure~S4. Loss-metric coupling on UPENN-GBM MRI (611
patients). Curves and markers are as in Figure 4 of the main text.}
\end{figure}

\begin{figure}
\hypertarget{fig:S5}{}\centering
\pandocbounded{\includegraphics[keepaspectratio,alt={\hyperlink{fig:S5}{Figure S5}. Loss-metric coupling on TCGA breast invasive carcinoma pathology (950 patients). Curves and markers are as in \hyperlink{fig:4}{Figure 4} of the main text.}]{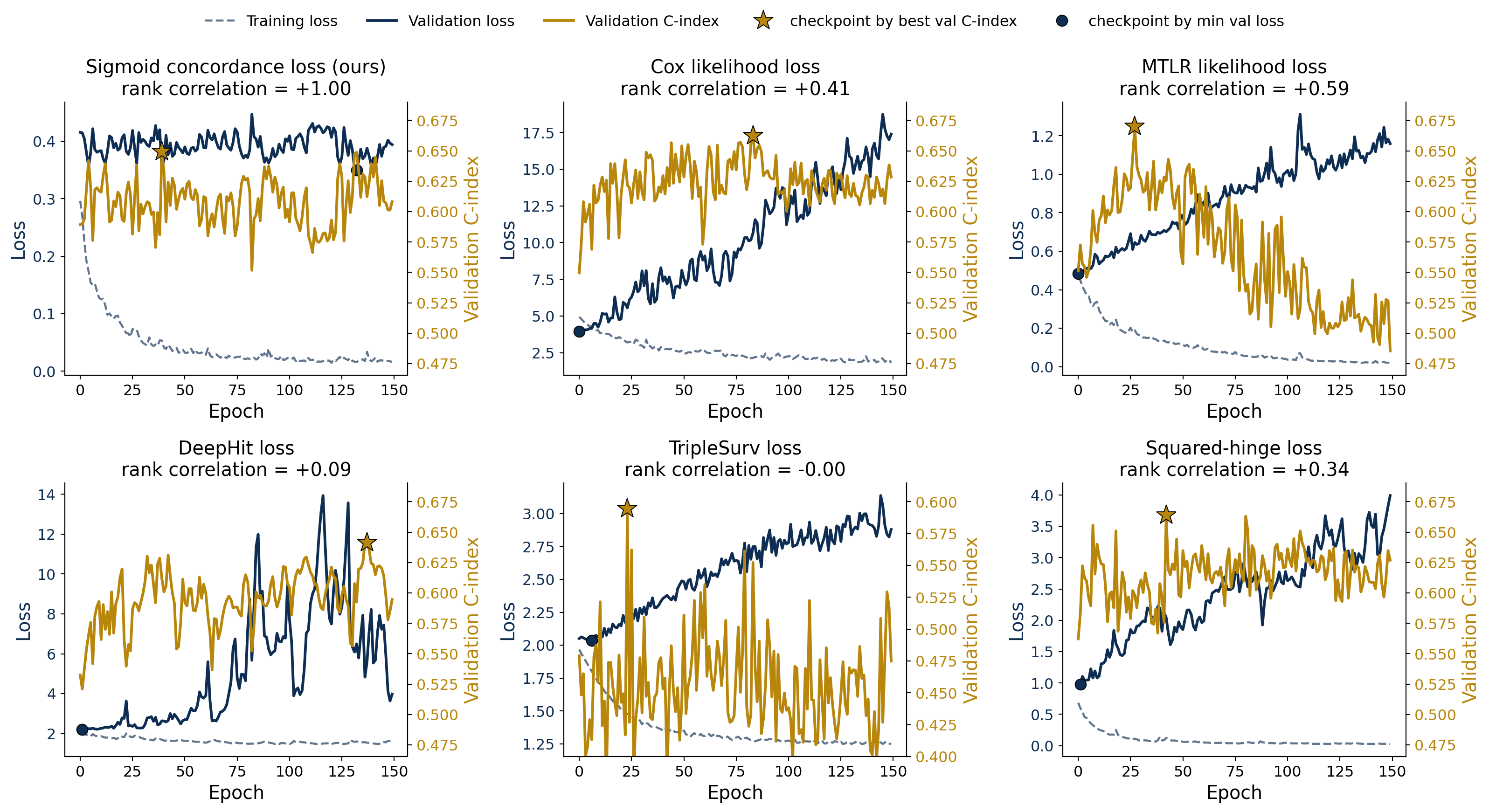}}
\caption{Figure~S5. Loss-metric coupling on TCGA breast invasive
carcinoma pathology (950 patients). Curves and markers are as in Figure
4 of the main text.}
\end{figure}

\begin{figure}
\hypertarget{fig:S6}{}\centering
\pandocbounded{\includegraphics[keepaspectratio,alt={\hyperlink{fig:S6}{Figure S6}. Loss-metric coupling on SUPPORT (9105 patients). Curves and markers are as in \hyperlink{fig:4}{Figure 4} of the main text.}]{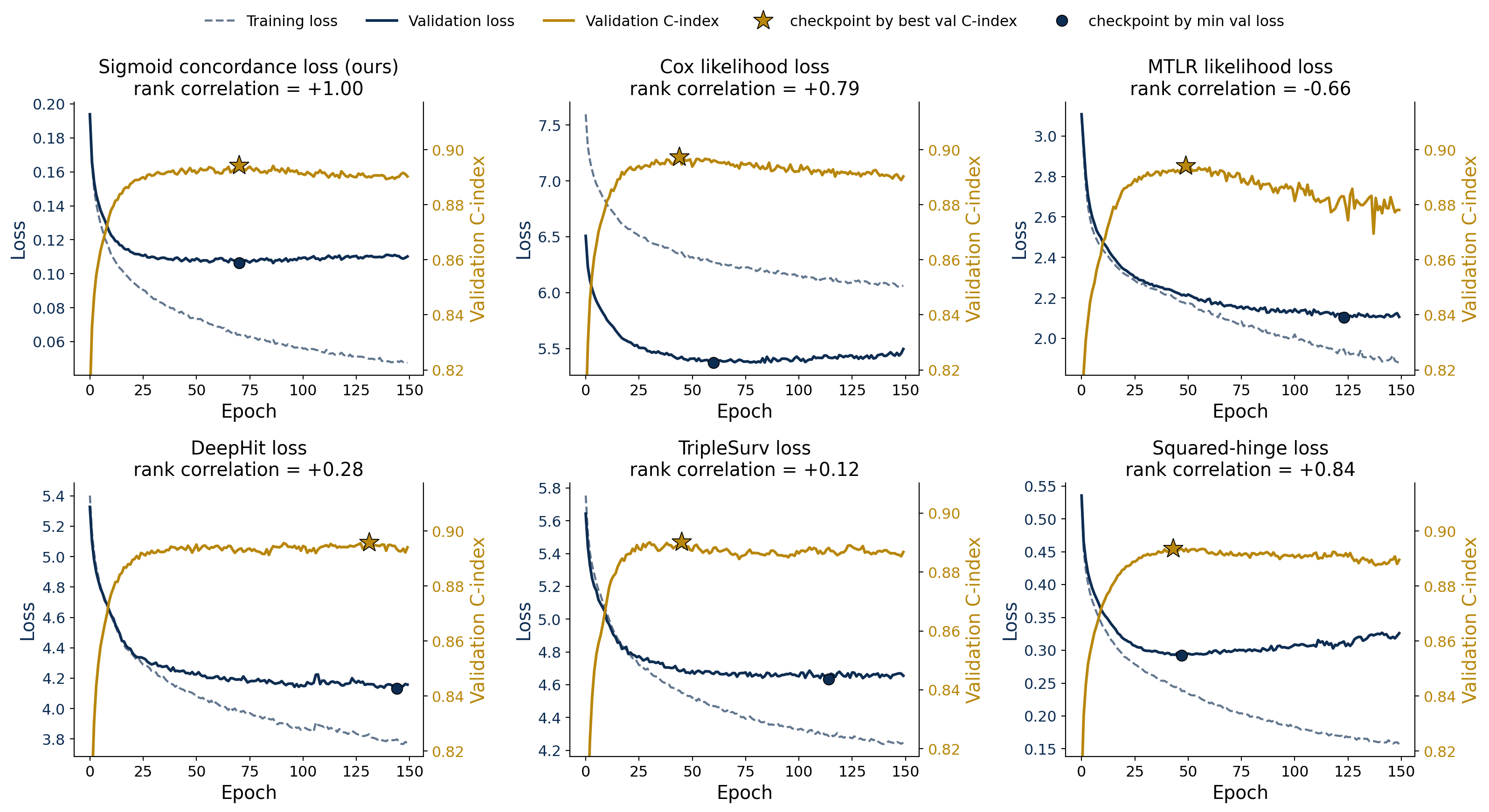}}
\caption{Figure~S6. Loss-metric coupling on SUPPORT (9105 patients).
Curves and markers are as in Figure 4 of the main text.}
\end{figure}

\begin{figure}
\hypertarget{fig:S7}{}\centering
\pandocbounded{\includegraphics[keepaspectratio,alt={\hyperlink{fig:S7}{Figure S7}. Loss-metric coupling on SEER (4024 patients). Curves and markers are as in \hyperlink{fig:4}{Figure 4} of the main text.}]{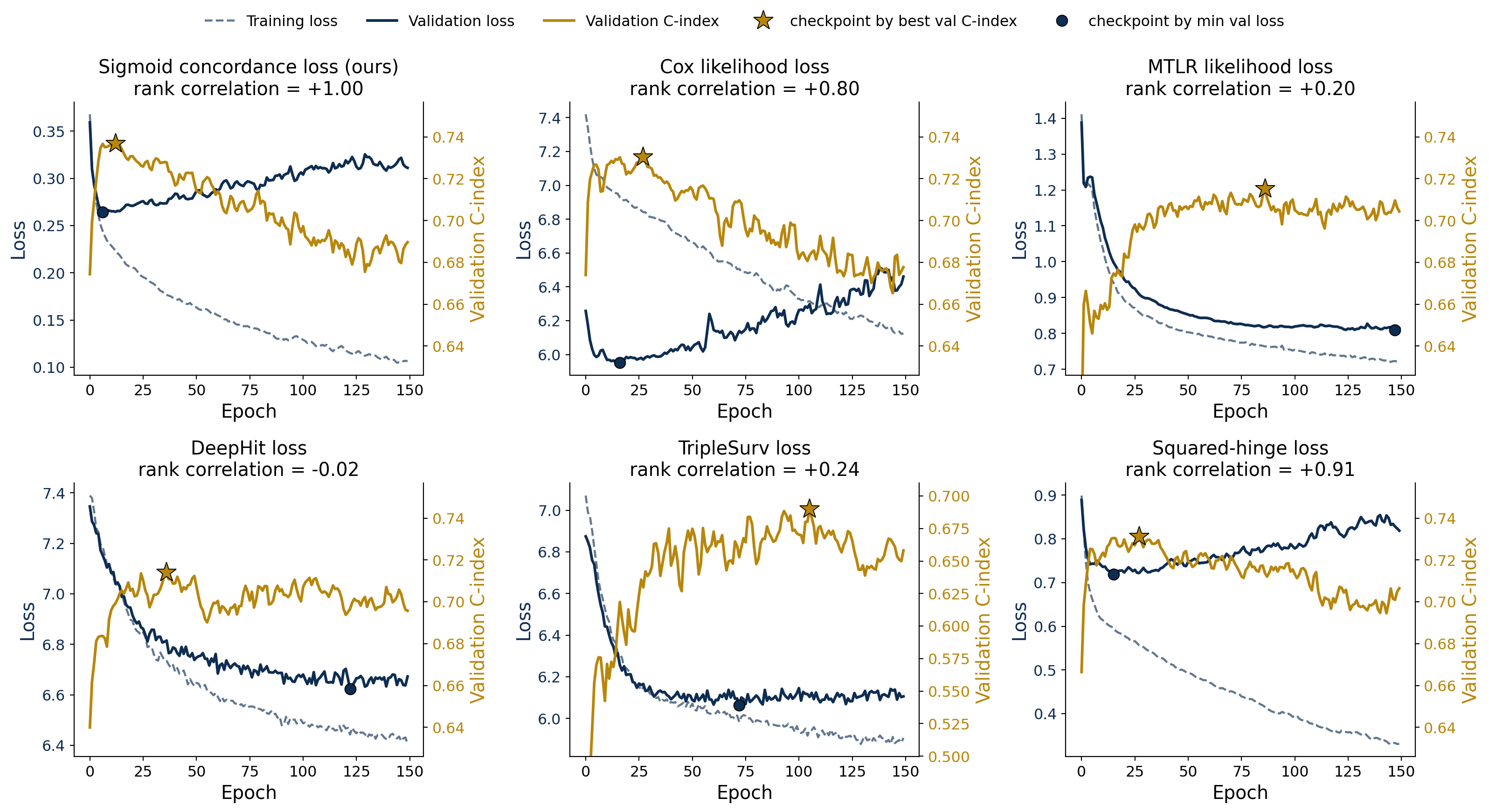}}
\caption{Figure~S7. Loss-metric coupling on SEER (4024 patients). Curves
and markers are as in Figure 4 of the main text.}
\end{figure}

\begin{figure}
\hypertarget{fig:S8}{}\centering
\pandocbounded{\includegraphics[keepaspectratio,alt={\hyperlink{fig:S8}{Figure S8}. Loss-metric coupling on FLCHAIN (7874 patients). Curves and markers are as in \hyperlink{fig:4}{Figure 4} of the main text.}]{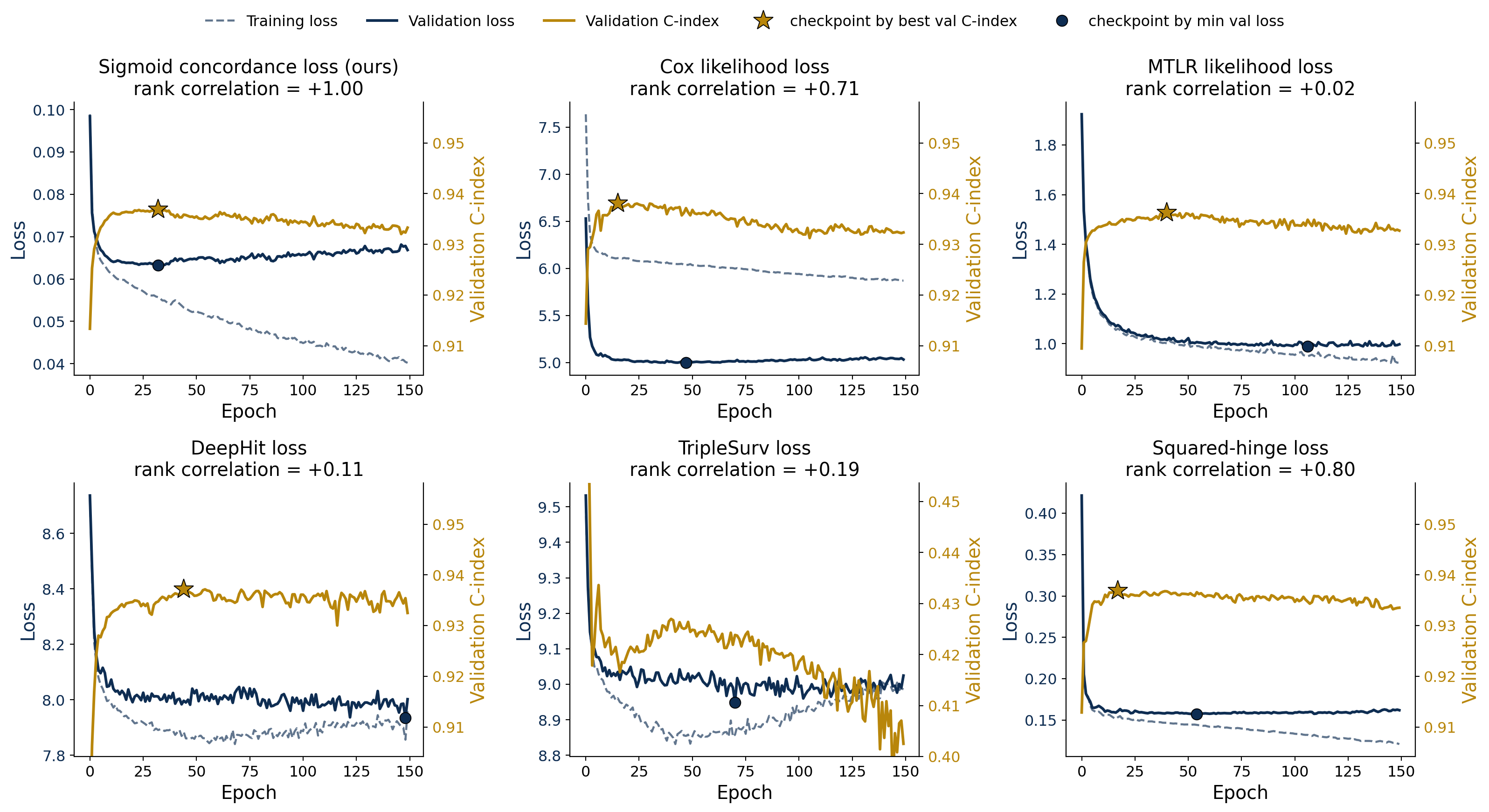}}
\caption{Figure~S8. Loss-metric coupling on FLCHAIN (7874 patients).
Curves and markers are as in Figure 4 of the main text.}
\end{figure}

\begin{figure}
\hypertarget{fig:S9}{}\centering
\pandocbounded{\includegraphics[keepaspectratio,alt={\hyperlink{fig:S9}{Figure S9}. Loss-metric coupling on GBSG (686 patients). Curves and markers are as in \hyperlink{fig:4}{Figure 4} of the main text.}]{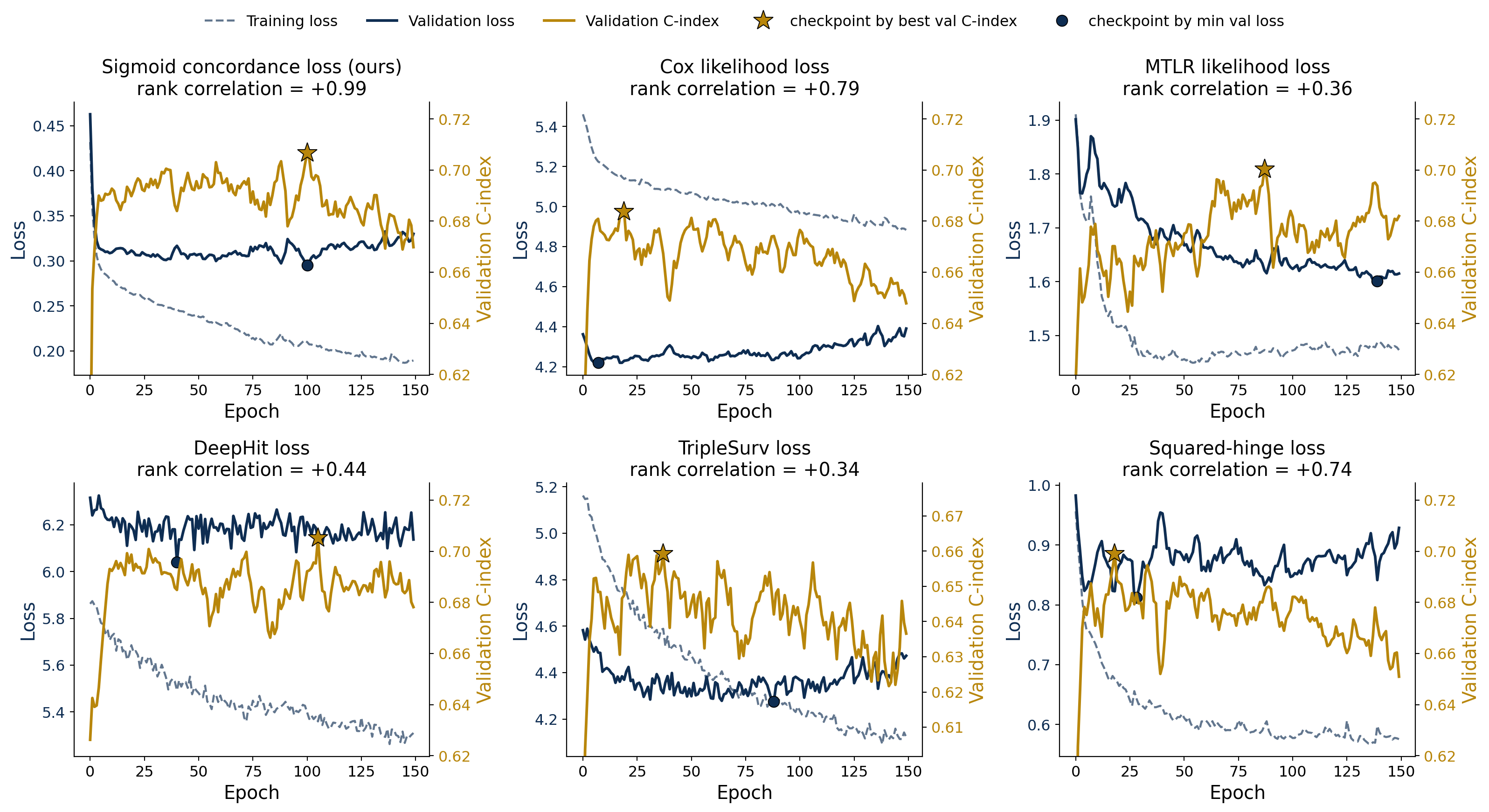}}
\caption{Figure~S9. Loss-metric coupling on GBSG (686 patients). Curves
and markers are as in Figure 4 of the main text.}
\end{figure}

\begin{figure}
\hypertarget{fig:S10}{}\centering
\pandocbounded{\includegraphics[keepaspectratio,alt={\hyperlink{fig:S10}{Figure S10}. Loss-metric coupling on WHAS500 (500 patients). Curves and markers are as in \hyperlink{fig:4}{Figure 4} of the main text.}]{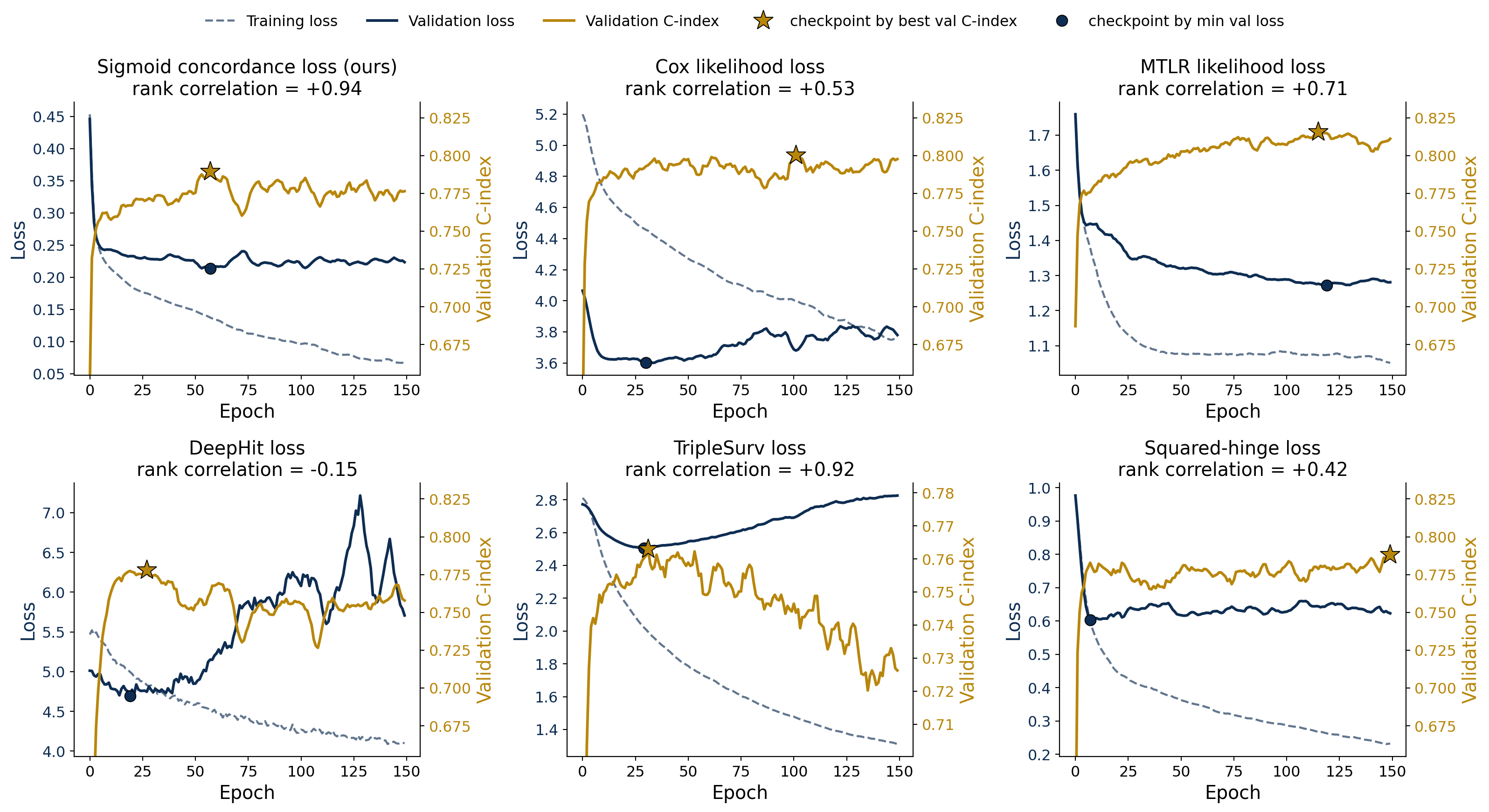}}
\caption{Figure~S10. Loss-metric coupling on WHAS500 (500 patients).
Curves and markers are as in Figure 4 of the main text.}
\end{figure}

\begin{figure}
\hypertarget{fig:S11}{}\centering
\pandocbounded{\includegraphics[keepaspectratio,alt={\hyperlink{fig:S11}{Figure S11}. Loss-metric coupling on PBC (418 patients). Curves and markers are as in \hyperlink{fig:4}{Figure 4} of the main text.}]{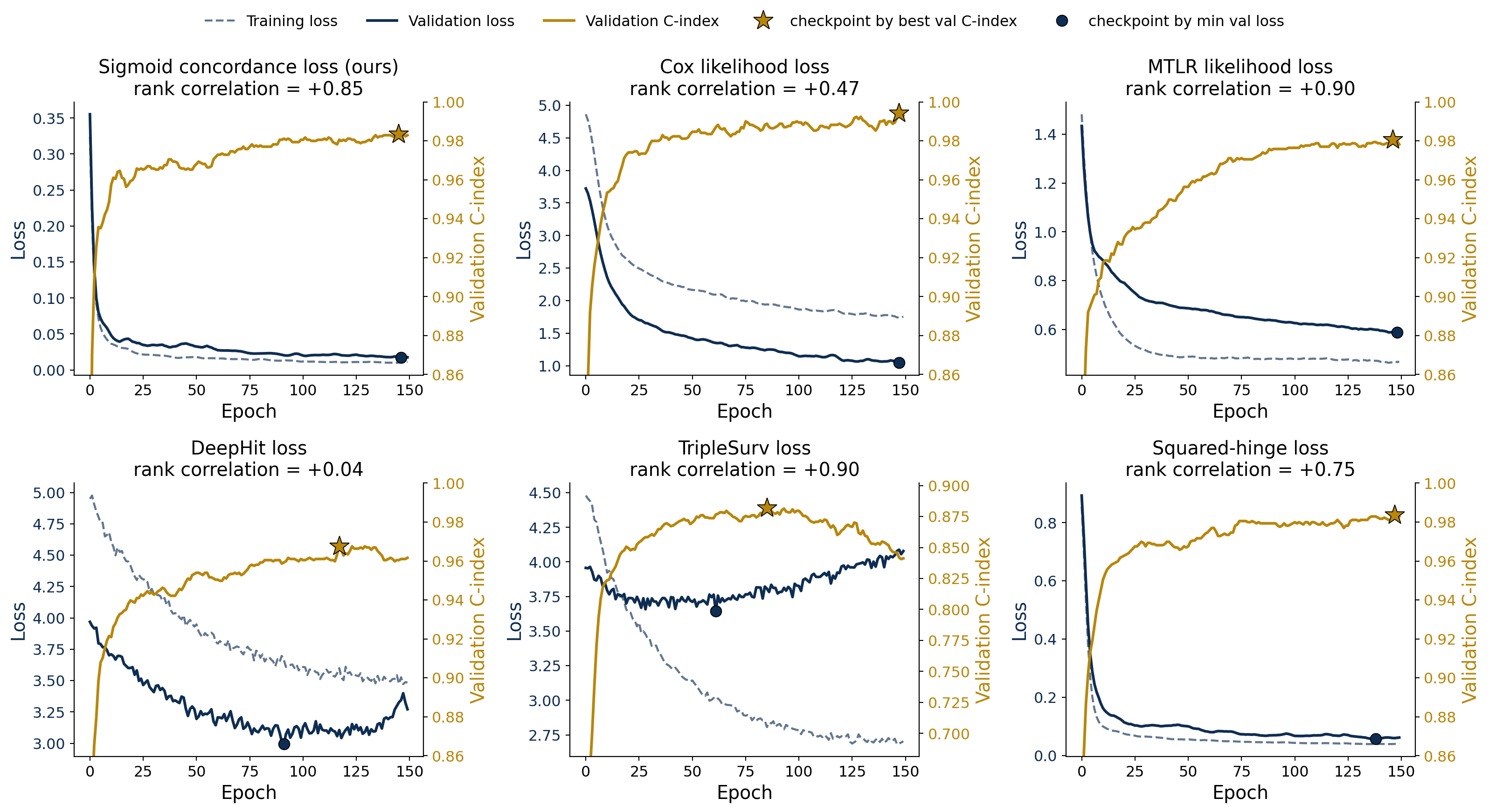}}
\caption{Figure~S11. Loss-metric coupling on PBC (418 patients). Curves
and markers are as in Figure 4 of the main text.}
\end{figure}

\end{document}